\documentclass[11pt, a4paper, logo, copyright]{googlecloud}

\pdftrailerid{redacted}

\makeatletter
\renewcommand\bibentry[1]{\nocitep{#1}{\frenchspacing\@nameuse{BR@r@#1\@extra@b@citeb}}}
\makeatother

\usepackage{kantlipsum, lipsum}
\usepackage{dsfont}

\usepackage[authoryear, sort&compress, round]{natbib}

\usepackage[utf8]{inputenc} %
\usepackage[T1]{fontenc}    %

\usepackage{microtype}
\usepackage{inconsolata}

\usepackage{booktabs}
\usepackage{graphicx}
\usepackage[most]{tcolorbox}
\usepackage{stackengine}
\usepackage{multirow}
\usepackage{subcaption} 
\usepackage{wrapfig}
\usepackage{hyperref}
\usepackage{makecell}
\usepackage{setspace}
\usepackage{booktabs} 
\usepackage{array}    
\usepackage{geometry} 

\usepackage{url}            
\usepackage{amsfonts}       
\usepackage{nicefrac}       
\usepackage{microtype}      
\usepackage{xcolor,colortbl}         
\usepackage{times}
\usepackage{latexsym}
\usepackage{multicol}
\usepackage{blindtext}
\usepackage{tabu}
\usepackage{amsmath, bm}

\usepackage{caption}
\usepackage[normalem]{ulem}
\usepackage{soul}
\usepackage{ragged2e}
\usepackage{pgffor}
\usepackage{textcomp}
\usepackage{amssymb}
\usepackage{pifont}
\usepackage{booktabs}
\usepackage{tabularx}
\usepackage{algorithm}
\usepackage{algpseudocode}

\definecolor{verylightgray}{gray}{0.9}

\definecolor{light_red}{rgb}{1.0, 0.6, 0.6}
\definecolor{light_green}{rgb}{0.56, 0.93, 0.56}

\newcommand{\plangen}[1]{PlanGEN}

\title{PlanGEN: A Multi-Agent Framework for Generating Planning and Reasoning Trajectories for Complex Problem Solving}

\correspondingauthor{Mihir Parmar \href{mailto:mihirparmar@asu.edu}{<mihirparmar@asu.edu>}, and Hamid Palangi \href{mailto:hamidpalangi@google.com}{<hamidpalangi@google.com>}}

\renewcommand{\today}{}

\author[1 2]{\fontsize{10.0pt}{10.0pt}\selectfont Mihir Parmar}
\author[1]{\fontsize{10.0pt}{10.0pt}\selectfont Xin Liu}
\author[1]{\fontsize{10.0pt}{10.0pt}\selectfont Palash Goyal}
\author[1]{\fontsize{10.0pt}{10.0pt}\selectfont Yanfei Chen}
\author[1]{\fontsize{10.0pt}{10.0pt}\selectfont Long Le}
\author[1]{\fontsize{10.0pt}{10.0pt}\selectfont Swaroop Mishra}
\author[1]{\fontsize{10.0pt}{10.0pt}\selectfont Hossein Mobahi}
\author[1]{\fontsize{10.0pt}{10.0pt}\selectfont Jindong Gu}
\author[1]{\fontsize{10.0pt}{10.0pt}\selectfont Zifeng Wang}
\author[1]{\fontsize{10.0pt}{10.0pt}\selectfont Hootan Nakhost}
\author[2]{\fontsize{10.0pt}{10.0pt}\selectfont Chitta Baral}
\author[1]{\fontsize{10.0pt}{10.0pt}\selectfont Chen-Yu Lee}
\author[1 $\clubsuit$]{\fontsize{10.0pt}{10.0pt}\selectfont Tomas Pfister}
\author[1 $\clubsuit$]{\fontsize{10.0pt}{10.0pt}\selectfont Hamid Palangi}

\affil[1]{\fontsize{9.0pt}{9.0pt}\selectfont Google}
\affil[2]{\fontsize{9.0pt}{9.0pt}\selectfont Arizona State University}

\begin{abstract}

Recent agent frameworks and inference-time algorithms often struggle with complex planning problems due to limitations in verifying generated plans or reasoning and varying complexity of instances within a single task. Many existing methods for these tasks either perform task-level verification without considering constraints or apply inference-time algorithms without adapting to instance-level complexity. To address these limitations, we propose \plangen{}, a model-agnostic and easily scalable agent framework with three key components: constraint, verification, and selection agents. Specifically, our approach proposes constraint-guided iterative verification to enhance performance of inference-time algorithms--Best of $\mathcal{N}$, Tree-of-Thought, and REBASE. In \plangen{} framework, the selection agent optimizes algorithm choice based on instance complexity, ensuring better adaptability to complex planning problems. Experimental results demonstrate significant improvements over the strongest baseline across multiple benchmarks, achieving state-of-the-art results on NATURAL PLAN ($\sim$8\%$\uparrow$), OlympiadBench ($\sim$4\%$\uparrow$), DocFinQA ($\sim$7\%$\uparrow$), and GPQA ($\sim$1\%$\uparrow$). Our key finding highlights that constraint-guided iterative verification improves inference-time algorithms, and adaptive selection further boosts performance on complex planning and reasoning problems.

\end{abstract}

\begin{document}

\maketitle

\section{Introduction}
\label{sec:introduction}

Effective planning is a crucial component for systems designed to solve complex real-world problems \citep{hao-etal-2023-reasoning, zhao-etal-2023-explicit, wang2024sibyl, jiao-etal-2024-learning, wang2025planning}. Traditional planning approaches, which rely heavily on template-based methods \citep{guan2023leveraging, valmeekam2024planbench, wang2024promptagent}, often lack generalizability and fail to capture the nuances of real-world tasks. In contrast, natural planning with LLMs offers a more promising direction, aligning better with real-world planning scenarios such as a trip or meeting planning \citep{zheng2024natural}. Furthermore, \citet{wang2025planning} shows that planning in natural language helps solve practical problems such as code generation. Thus, we aim to enhance LLMs' ability to generate effective natural plans and demonstrate their usefulness in solving downstream reasoning tasks within the scientific and financial domains. For the scope of this study, ``planning'' refers to the ability to decompose tasks and reason strategically to achieve solutions.

In recent years, LLM agents have shown impressive abilities to solve complex reasoning problems \citep{yao2023react, xiao2024chainofexperts, wang2024survey}. Orthogonal to this exploration, scaling a search space during inference-time (i.e., test-time scaling) \citep{snell2024scaling, welleck2024decoding} has gained popularity in tackling difficult problems such as mathematical reasoning \citep{zhang2024accessing} and code generation \citep{wang2025planning}. Despite the success of these frameworks, we hypothesize that they often struggle with complex planning problems due to the lack of better verification module, and a failure to account for instance-level complexity across single-task. Furthermore, although some initial explorations exist \citep{bohnet2024exploring, lee2025evolving}, effectiveness of these frameworks for natural planning is under-explored (extended related work is presented in App. \ref{sec:related_works}). Motivated by these, we proposed \plangen{}, a model-agnostic, easily scalable, multi-agent framework for effective natural plan generation. 

\begin{figure*}
    \centering
    \includegraphics[width=0.8\textwidth]{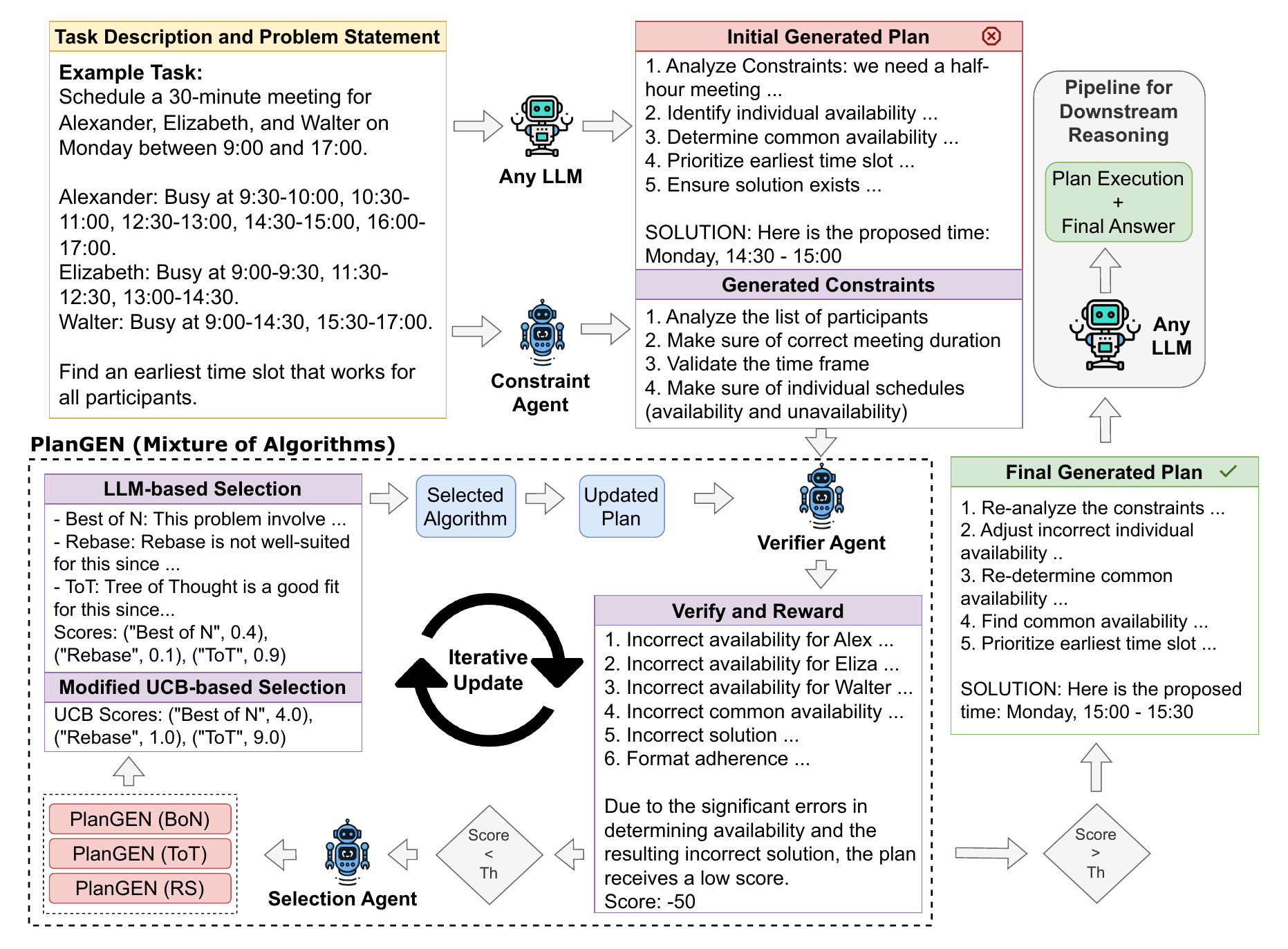}
    \caption{Schematic representation of \plangen{} (Mixture of Algorithms). An initial plan and constraints guide iterative plan refinement. The verification agent provides reward scores for plan quality, and the selection agent chooses inference algorithms until the highest-reward plan is found and used for downstream reasoning (if needed). UCB: Upper Confidence Bound, BoN: Best of $\mathcal{N}$, ToT: Tree-of-Thought, RS: REBASE.}
    \label{fig:teaser}
\end{figure*}

\plangen{} consists of three specialized agents: \textit{constraint agent}, \textit{verification agent}, and \textit{selection agent}. The constraint agent extracts instance-specific constraints (e.g., budget, concepts, rules, etc.); the verification agent evaluates plan quality and provides a reward score considering the constraints; and the selection agent dynamically chooses the best inference algorithm using an improved Upper Confidence Bound (UCB) formula \citep{han2024ucb} for instance of different complexity. We explore popular and widely used three inference algorithms within \plangen{}: Best of $\mathcal{N}$ \citep{brown2024large}, Tree-of-Thought (ToT) \citep{yao2024tree}, and REward-BAlanced SEarch (REBASE) \citep{wu2024empirical}. We combine our agents with these algorithms, yielding four frameworks: (1) \plangen{} (Best of $\mathcal{N}$), (2) \plangen{} (ToT), (3) \plangen{} (REBASE), and (4) \plangen{} (Mixture of Algorithms).  In \plangen{}, we use ``Multi-Agent'' approach which signifies using the constraint and verification agents for the first three approaches, and all three agents for the ``Mixture of Algorithms'' (Figure \ref{fig:teaser}). Figure \ref{fig:teaser} shows example from NATURAL PLAN (Calendar scheduling), and App. \ref{app:examples} provides more examples.

To evaluate \plangen{}, we perform all experiments using Gemini-1.5-Pro \citep{team2024gemini} as underlying model. We further present case-study on Gemini-2.0-Flash, and GPT-4o \citep{hurst2024gpt} to show the model-agnostic nature. We evaluate natural language planning ability on NATURAL PLAN \citep{zheng2024natural}, scientific/mathematical reasoning on GPQA \citep{rein2024gpqa} and OlympiadBench \citep{he-etal-2024-olympiadbench}, and financial reasoning on DocFinQA \citep{reddy-etal-2024-docfinqa}. Performance is compared against Zero-shot Chain-of-Thought (CoT) and a vanilla multi-agent baselines. We achieve state-of-the-art results on NATURAL PLAN ($\sim$8\%$\uparrow$ average across all categories), OlympiadBench (text-only) ($\sim$5\%$\uparrow$ on MATH, $\sim$4\%$\uparrow$ on PHYSICS), and DocFinQA ($\sim$7\%$\uparrow$). On GPQA, we outperform Gemini-1.5-Pro ($\sim$13\%$\uparrow$), GPT-4o ($\sim$12\%$\uparrow$), and Claude-3.5-Opus ($\sim$9\%$\uparrow$), while achieving competitive performance compared to the vanilla multi-agent baseline ($\sim$1\%$\uparrow$). Further analysis reveals that the simplest method (i.e., \plangen{} (Best of $\mathcal{N}$)) achieves the best performance on NATURAL PLAN (Figure \ref{fig:main_results}). \plangen{} (Mixture of Algorithms) achieves the best performance for complex problems (Figure \ref{fig:np_cal_analysis}) including GPQA, and OlympiadBench(MATH). We further conduct a thorough analysis of the results which reveals several important findings. In summary, our contributions are: (1) \plangen{}, a novel, model-agnostic, and scalable multi-agent framework for enhancing LLM natural planning; (2) SOTA results on several complex planning and reasoning benchmarks; and (3) a novel approach to constraint-based verification and instance-level complexity-based inference algorithm selection.

\section{\plangen{}}
\label{sec:method}


\subsection{Proposed LLM Agents}

\plangen{} comprises three specialized LLM agents: a \textit{constraint agent}, a \textit{verification agent}, and a \textit{selection agent}. Each agent utilizes an off-the-shelf LLM (e.g., Gemini, GPT) which is equipped with task-specific prompts for efficient performance.

\subsubsection{Constraint Agent}
\label{subsec:constraint}

We define ``constraints'' as the criteria necessary for verifying solutions to planning problems. These criteria are inherently instance-specific. For instance, in the calendar scheduling from NATURAL PLAN, relevant constraints include `individual schedules', `availabilities', and `preferences'. In a scientific reasoning problems from GPQA, constraints might be the `concepts used', `calculation correctness', and `formula selection'. We argue that careful extraction of instance-specific constraints is critical for successful verification. The constraint agent serves as a preprocessing component in the framework, designed to extract instance-specific constraints from the problem description. By analyzing the input problem, this agent identifies all possible critical constraints that are required for generated plan verification. The extracted constraints provide a foundation for verifying plans to improve the overall relevance and quality of the planning process. The prompt used by the constraint agent enables it to systematically identify constraints by asking the underlying LLM to focus on specific aspects of the problem description. This ensures that no critical information is overlooked and that the resulting constraints are comprehensive. Prompts used by the constraint agent and examples of generated constraints are provided in App. \ref{app:llm_agents} and App. \ref{app:examples}, respectively.

\subsubsection{Verification Agent}
\label{subsec:verification}

The verification agent plays a critical role in the framework by assessing the quality of generated plans based on constraints generated by the constraint agent. This agent ensures that plans are aligned with task objectives, adhere to constraints, and progress logically toward a correct and complete solution. The verification agent has two key components: (i) feedback generation, and (ii) numerical reward score generation based on feedback. Verification prompts and examples of verification are provided in App. \ref{app:llm_agents} and App. \ref{app:examples}, respectively.

\paragraph{Feedback Generation}
While verifying each generated plan against different constraints, the verification agent generates detailed natural language reasoning regarding plan quality. We consider this explanation as ``feedback'', offering interpretability and actionable next step towards improvement.

\paragraph{Numerical Reward Generation}
Motivated by \citet{zhang2024accessing}, we instruct the agent to evaluate the plan against various constraints and assign a reward score on a scale of $-100$ to $100$. The scoring mechanism is designed to enforce strict quality standards, with a threshold (e.g., a score of $95$ or higher) indicating a verified, high-quality plan.

\subsubsection{Selection Agent}
\label{subsec:selection}

The selection agent dynamically determines the most suitable inference algorithm for solving a given problem instance based on its complexity. It leverages a combination of historical performance; diversity, and recovery scores; and guidance from a LLM to adaptively select the best algorithm for the current instance. To create the selection agent, we utilize a modified Upper Confidence Bound (UCB) policy. The policy combines multiple factors, including normalized rewards, exploration bonuses, diversity adjustments, and recovery scores. Additionally, the agent incorporates LLM-guided priors, which provide algorithm suitability scores based on the problem statement, task requirements, and previous plan (if available). These priors enable the agent to align its selections with the input instance complexity and corresponding constraints, improving the relevance of the chosen algorithm.


\paragraph{Modified UCB Policy} equation combines several terms to balance exploitation, and exploration when selecting the best algorithm for given task instance. To modify UCB, we first conducted a preliminary ablation study, presented in App. \ref{app:llm_agents}.

\begin{align*}
\mathrm{UCB}(a) = \frac{R(a)}{N(a)} + \sqrt{\frac{2 \log(T + 1)}{N(a)}} + \lambda_{\mathrm{prior}} \cdot \mathrm{Prior}(a) + \frac{\alpha_{\mathrm{diversity}}}{N(a) + 1} + \alpha_{\mathrm{recovery}} \cdot S_{\mathrm{recovery}}(a)
\end{align*}

All terms in equation given above are calculated across one evaluation run. Here, the cost of calculation is negligible since it only utilizes reward values from previous runs, but only one LLM call require to get score for \text{Prior}(a). The first term, \(\frac{R(a)}{N(a)}\), represents the average reward for algorithm \(a\), where \(R(a)\) is the total reward accumulated by the algorithm, and \(N(a)\) is the number of times the algorithm has been selected. This term ensures that algorithms with higher historical performance are prioritized. The second term, \(\sqrt{\frac{2 \log(T + 1)}{N(a)}}\), serves as the exploration component, encouraging the selection of algorithms with fewer trials, denoted as $T$. This term ensures that under-explored options are adequately evaluated. Furthermore, \(\lambda_{\text{prior}} \cdot \text{Prior}(a)\), which leverages LLM-guided priors to align algorithm selection with the instance-specific complexity. Here, \(\lambda_{\text{prior}}\) is a dynamically decaying weight defined as \(\frac{\lambda_{\text{prior}}}{1 + T}\), where \(T\) represents the total number of trials. This decay gradually shifts the focus from initial priors to historical performance as trials progress. The diversity bonus, \(\frac{\alpha_{\text{diversity}}}{N(a) + 1}\), penalizes overused algorithms, ensuring balanced exploration across all options. Finally, the recovery term, \(\alpha_{\text{recovery}} \cdot S_{\text{recovery}}(a)\), rewards algorithms that recover effectively from failures, with \(S_{\text{recovery}}(a)\) representing the recovery score for algorithm \(a\).

\paragraph{Selection Process}
This process begins by initializing algorithm-specific variables, such as accumulated rewards, selection counts, and failure counts. Further details on the algorithm can be found in Algorithm \ref{algo:selection} (App. \ref{app:llm_agents}). The agent then incorporates LLM-guided priors to generate suitability scores for the algorithms based on the problem statement and any provided feedback. These priors are derived from a LLM (prompt for this given in App. \ref{app:llm_agents}), and serve as initial estimates to adjust the UCB \citep{han2024ucb} values.




\subsection{Proposed Frameworks}
\label{subsec:frameworks}

Within \plangen{}, we propose four different frameworks: (1) \plangen{} (Best of $\mathcal{N}$) (Figure \ref{fig:bon}), (2) \plangen{} (ToT) (Figure \ref{fig:tot}), and (3) \plangen{} (REBASE) (Figure \ref{fig:rebase}), and (4) \plangen{} (Mixture of Algorithms) (Figure \ref{fig:teaser}). 


\subsubsection{\plangen{} (Best of $\mathcal{N}$)}

\begin{wrapfigure}{r}{0.45\textwidth}
    \centering
    \vspace{-20mm}
    \includegraphics[width=0.8\linewidth]{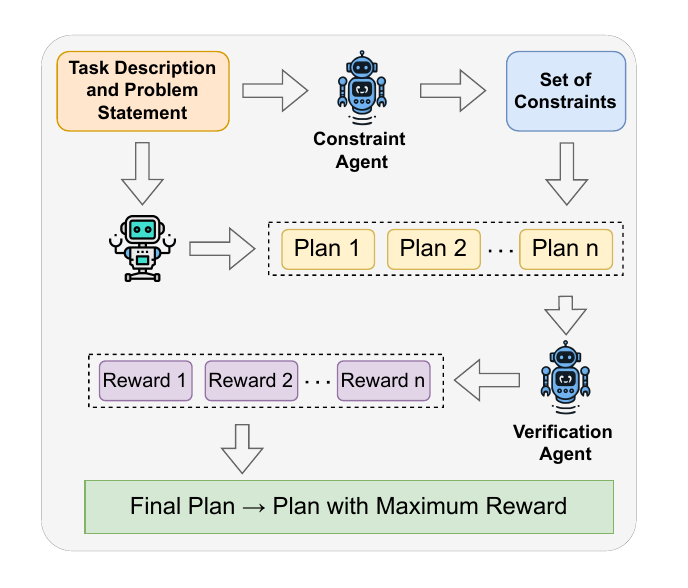}
    \vspace{-3mm}
    \caption{Schematic representation of \plangen{} (Best of $\mathcal{N}$) (BoN).}
    \label{fig:bon}
\end{wrapfigure}Motivated by \citet{brown2024large}, we adapted the Best of $\mathcal{N}$ algorithm and modified it using our constraint and verification agents as illustrated in Figure \ref{fig:bon}.  The framework generates $\mathcal{N}$ candidate plans (Plan 1, Plan 2, ..., Plan n), and each plan is assessed by a verification agent based on a set of constraints. Then, a corresponding reward (Reward 1, Reward 2, ..., Reward n) gets assigned by the verification agent. Finally, the plan with the maximum reward is chosen, guaranteeing an optimal solution that best satisfies the problem constraints.

\subsubsection{\plangen{} (ToT)}

\begin{wrapfigure}{r}{0.45\textwidth}
    \centering
    \vspace{-12mm}
    \includegraphics[width=0.8\linewidth]{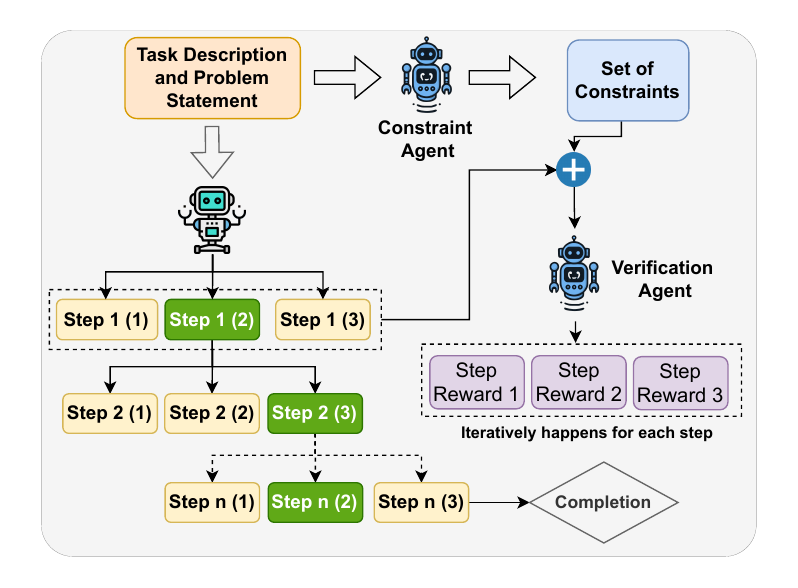}
    \vspace{-3mm}
    \caption{Schematic representation of \plangen{} (ToT). Highest-reward steps are highlighted in green.}
    \label{fig:tot}
\end{wrapfigure}ToT algorithm has been studied in detail for solving many complex problems \citep{yao2024tree}. As shown in Figure \ref{fig:tot}, we modify the ToT algorithm with our constraint and verification agents. The method begins by initializing a root node that represents the problem and generating multiple potential next steps, creating a tree-like structure. The generated steps are verified using a verification agent which assigns reward scores based on a set of constraints. The iterative process involves evaluating all possible steps at a given depth, selecting the most promising path based on reward scores, and expanding it further by generating new steps. This process continues until a valid solution is identified or a pre-defined limit on iterations is reached. Further details on various prompts for the ToT are presented in App. \ref{app:frameworks}.


\subsubsection{\plangen{} (REBASE)}

\begin{wrapfigure}{r}{0.45\textwidth}
    \centering
    \vspace{-12mm}
    \includegraphics[width=0.8\linewidth]{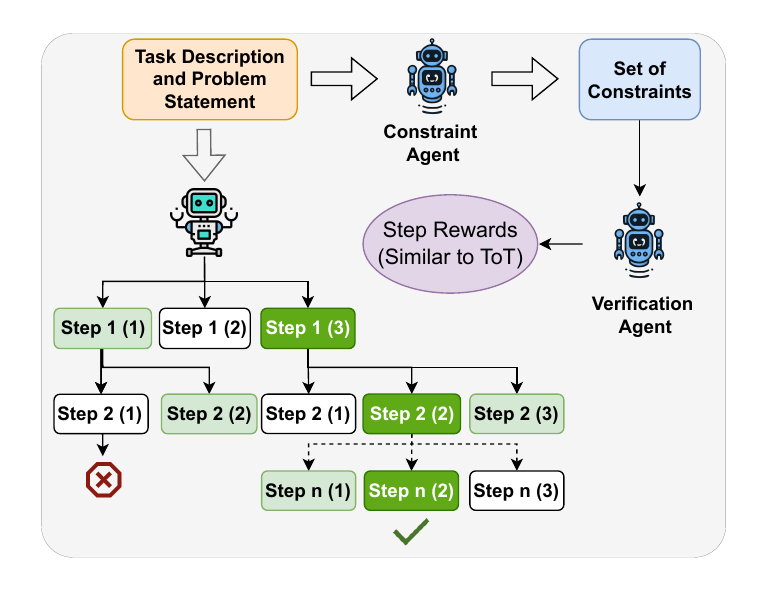}
    \vspace{-3mm}
    \caption{Schematic representation of \plangen{} (REBASE). Green shading indicates step reward (darker $=$ higher).  Darker steps prioritized for exploration.}
    \label{fig:rebase}
\end{wrapfigure}

The REBASE tree search method inherits the exploitation and pruning properties of tree search and is well-studied for mathematical reasoning \citep{wu2024empirical}. As shown in Figure \ref{fig:rebase}, the framework incorporates a dynamic selection and expansion strategy to iteratively refine solutions. At each depth of the tree, candidate nodes are ranked based on their assigned reward scores (obtained using a verification agent), ensuring that the most promising candidates are explored first. Even steps with lower rewards are considered but with a reducing number of children, meaning that their exploration depth is limited. This hierarchical pruning helps maintain efficiency, thereby reducing unnecessary exploration of weaker nodes. This process continues until either a valid, complete solution is found or a predefined depth or width limit is reached. Also, there is a completion check similar to ToT which identifies nodes that represent complete solutions, enabling REBASE to terminate early once a satisfactory outcome is identified. App. \ref{app:frameworks} provides further details on prompts for the REBASE.


\subsubsection{\plangen{} (Mixture of Algorithms)}

The Mixture of Algorithms framework (Figure \ref{fig:teaser}) introduces a selection agent (\textsection \ref{subsec:selection}) which dynamically selects the best possible inference-time algorithms proposed in the above sections based on instance-level complexity. The framework operates in a modular and iterative manner, ensuring adaptability in addressing planning and reasoning problems with different complexity effectively.


\paragraph{Orchestration} 
The process begins with generating an initial plan using LLM based on the task description and problem statement. Along with this, the constraint agent (\textsection \ref{subsec:constraint}) is employed to generate an instance-specific set of constraints. Based on the constraints, the verification agent (\textsection \ref{subsec:verification}) evaluates the quality of the initial plan and provides a reward score (indicated as `Score' in Figure \ref{fig:teaser}). If the initial plan meets the required threshold (denoted $T_h$), it is acceptable as the ``Final Plan''. Otherwise, the iterative refinement process begins.

\paragraph{Iterative Refinement}
The refinement loop is driven by a suite of inference algorithms as shown in Figure \ref{fig:teaser}. During this iterative refinement, the selection agent (\textsection \ref{subsec:selection}) determines the most suitable algorithm based on the instance-specific complexity and historical UCB values. The selected algorithm produces an updated plan, which is then re-evaluated by the verification agent. To ensure continual improvement, the framework incorporates feedback generated by a verification agent that provides guidance, and this feedback loop enables the system to refine the plan incrementally.

\section{Experiments and Results}
\label{sec:exp_and_res}

\subsection{Experimental Setup}


\paragraph{Datasets} 
To demonstrate improvement in natural planning abilities, we utilize the NATURAL PLAN \citep{zheng2024natural}. After improving the planning, we show that this significantly enhances the reasoning capabilities of LLMs on two benchmarks: GPQA \citep{rein2024gpqa} and OlympiadBench (text-only) \citep{he-etal-2024-olympiadbench}. Additionally, we show that \plangen{} improves performance on a domain-specific dataset, DocFinQA \citep{reddy-etal-2024-docfinqa}. Further details are presented in App. \ref{app:experiments}.

\begin{figure}[ht]
    \centering
    \begin{subfigure}[b]{0.49\textwidth}
        \centering
        \includegraphics[width=\textwidth]{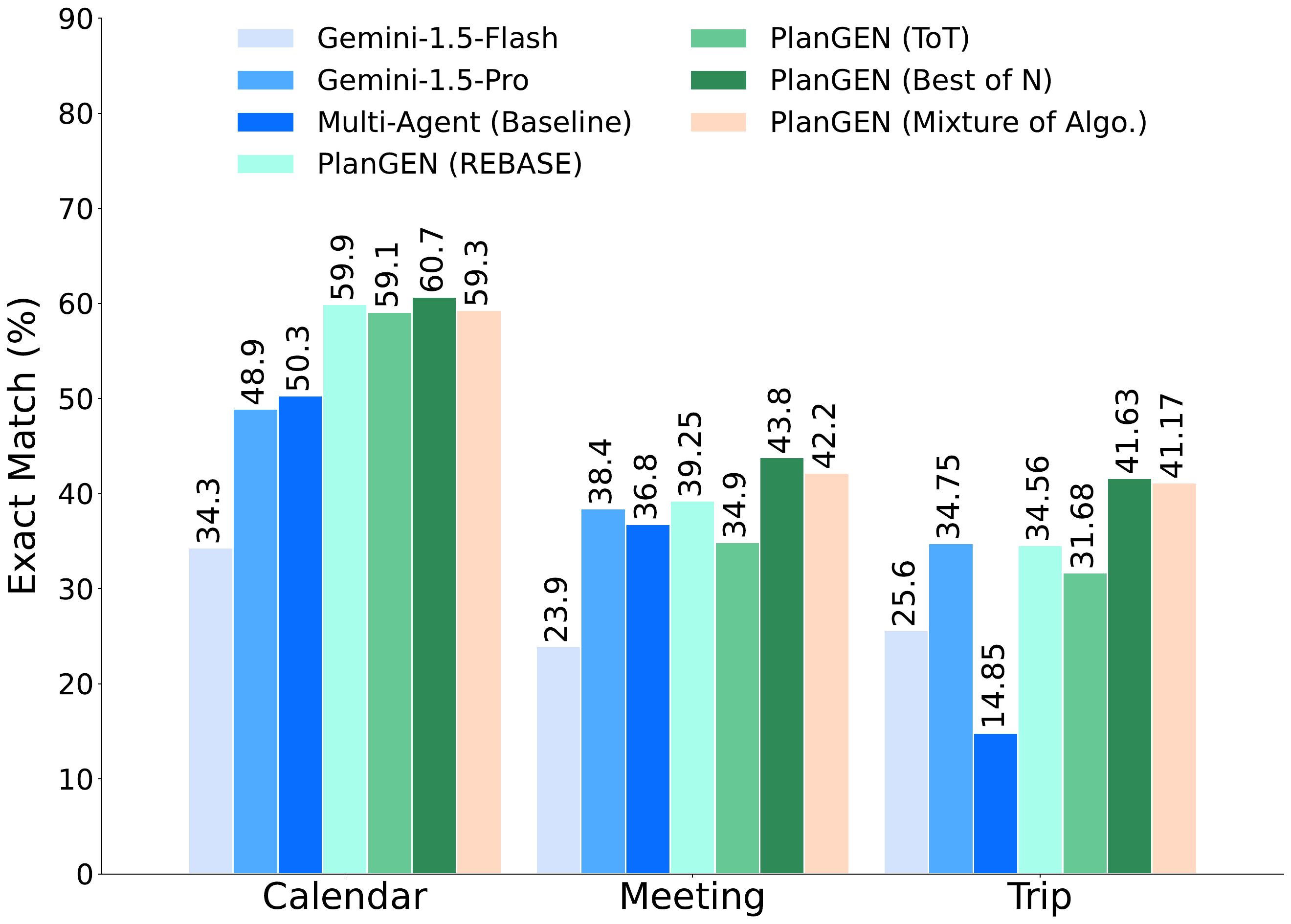} 
        \caption{\centering NATURAL PLAN}
        \label{subfig:natural_plan}
    \end{subfigure}
    \hfill
    \begin{subfigure}[b]{0.49\textwidth}
        \centering
        \includegraphics[width=\textwidth]{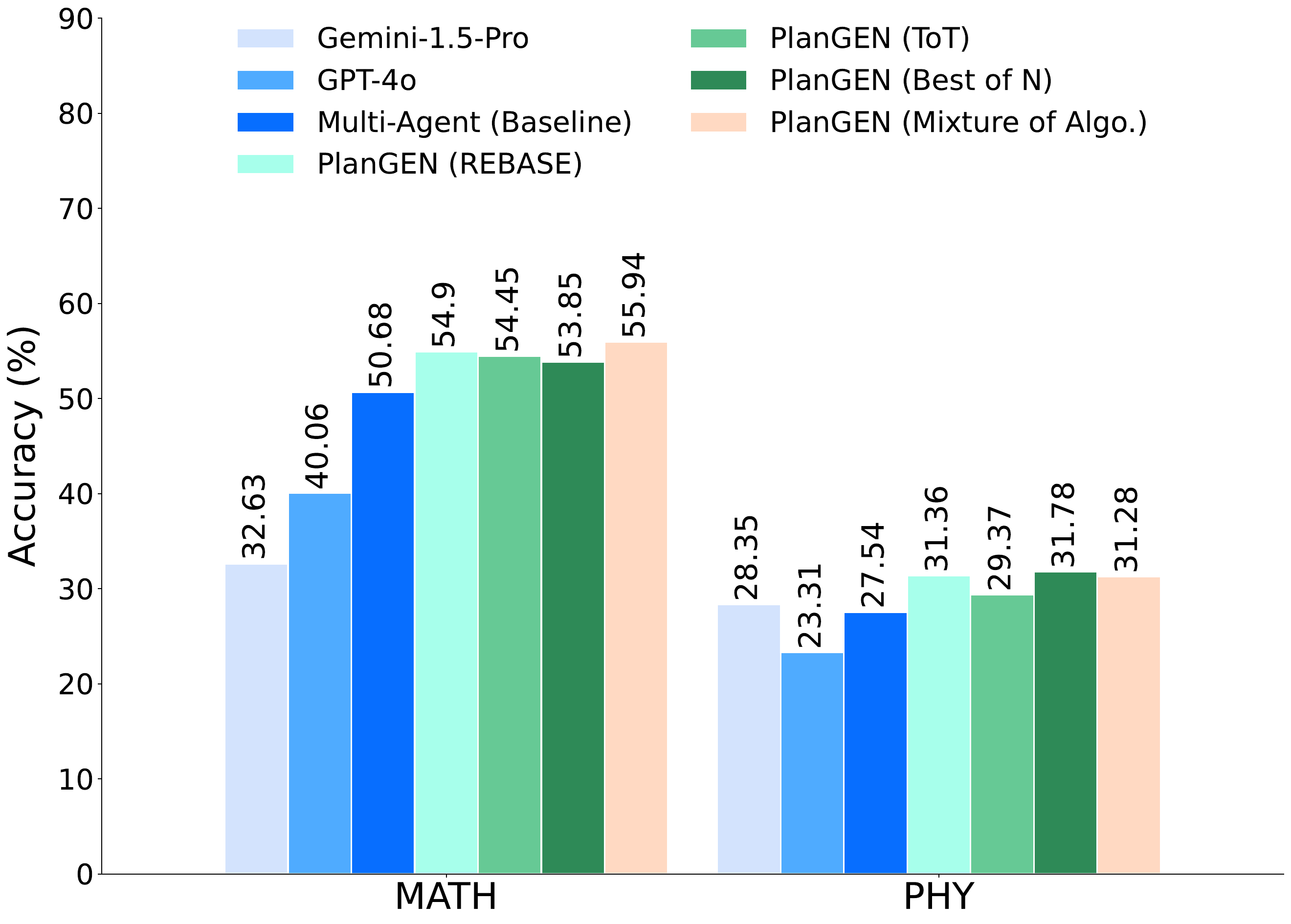} 
        \caption{\centering OlympiadBench}
        \label{subfig:olympiad}
    \end{subfigure}
    
    \vspace{\baselineskip} 
    
    \begin{subfigure}[b]{0.49\textwidth}
        \centering
        \includegraphics[width=\textwidth]{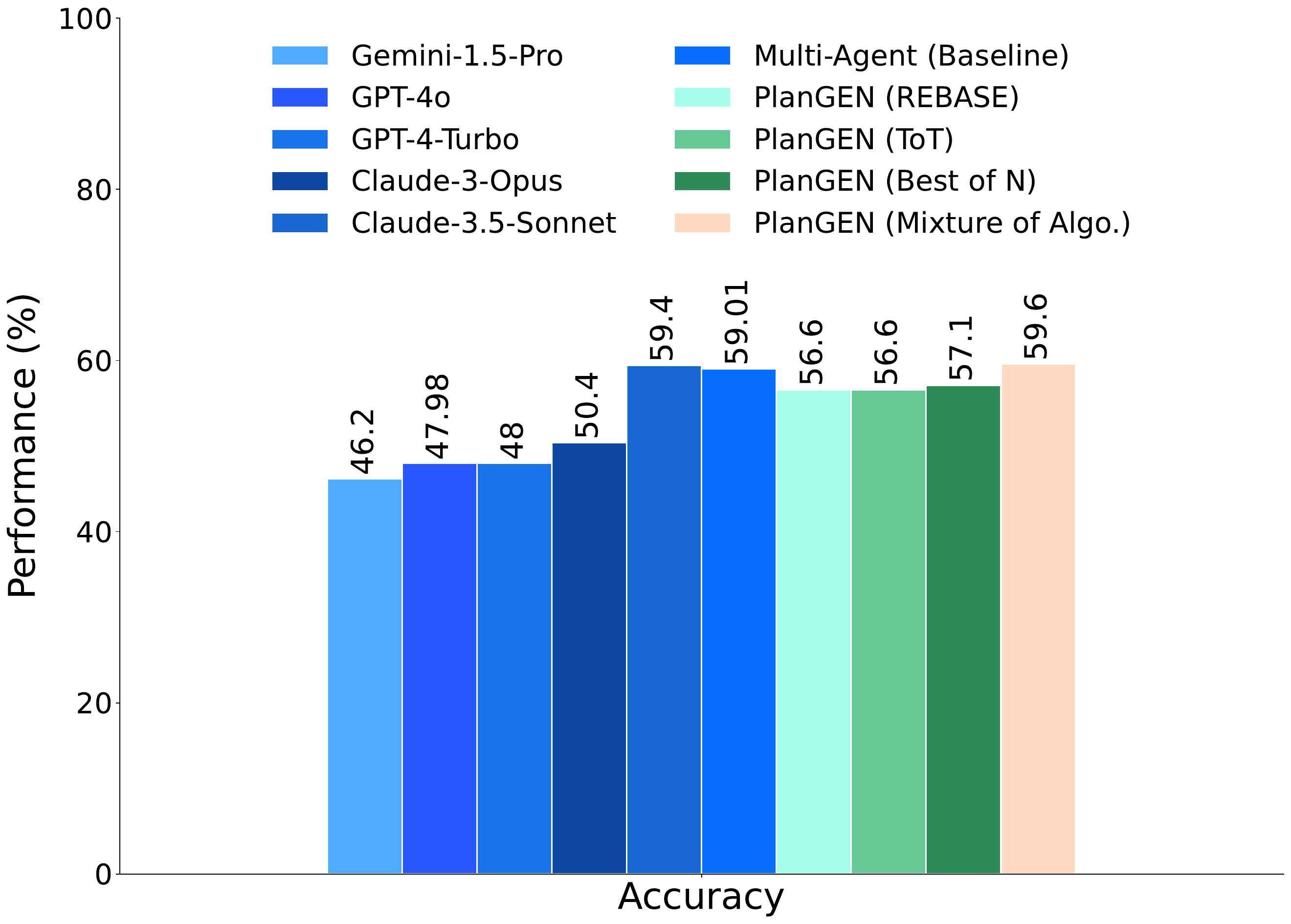} 
        \caption{\centering GPQA}
        \label{subfig:gpqa}
    \end{subfigure}
    \hfill
    \begin{subfigure}[b]{0.49\textwidth}
        \centering
        \includegraphics[width=\textwidth]{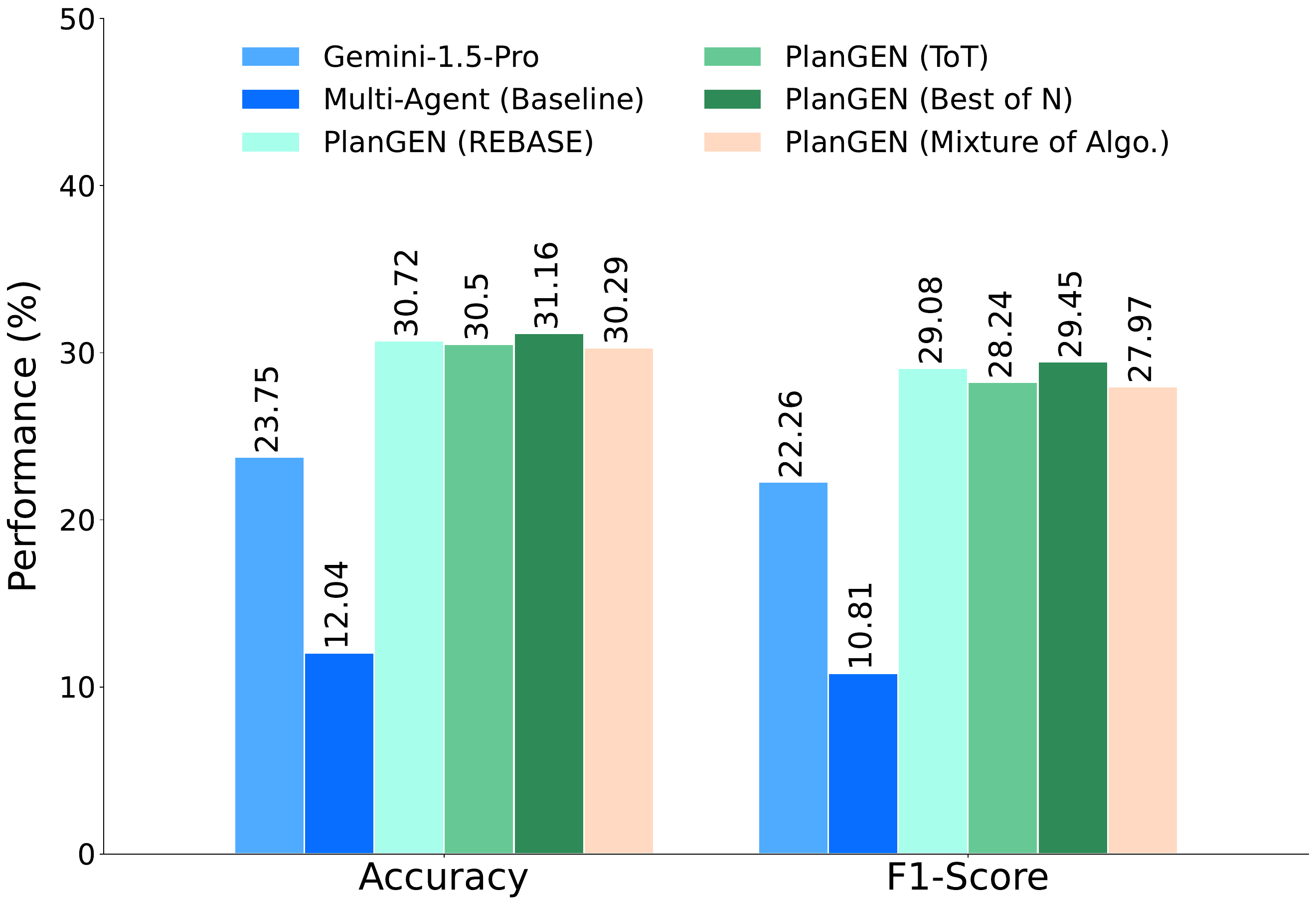} 
        \caption{\centering DocFinQA}
        \label{subfig:docfinqa}
    \end{subfigure}
    
    \caption{Performance comparison of the proposed multi-agent frameworks against baselines across four benchmarks. All experiments are conducted using Gemini-1.5-Pro. Algo: Algorithms, MATH: Mathematics, PHY: Physics.}
    \label{fig:main_results}
\end{figure}

\paragraph{Baselines and Our Frameworks} 
We develop two baselines for comparison with our frameworks: (i) Zero-shot CoT \citep{zero_shot_cot} and (ii) a Vanilla Multi-Agent Baseline. In the Zero-shot CoT, we provide an input prompt to the model, which generates outputs in the form of <CoT reasoning, Answer>. For the ``Multi-Agent Baseline'', the same model is called iteratively across multiple iterations. The system repeatedly refines its outputs through feedback loops, where the feedback is generated based on a self-reflective prompt (App. \ref{app:experiments}) designed to improve reasoning. We evaluate all proposed frameworks (\textsection \ref{subsec:frameworks}) on all benchmarks. For reasoning tasks, we use a two-stage approach: (1) generating an optimized plan using our frameworks, and (2) executing the plan to produce the final answer (Figure \ref{fig:teaser}). App. \ref{app:experiments} presents further details on model selection, metrics, and experiment hyper-parameters including the hyper-parameter choices for inference-time algorithms.


\subsection{Main Results}

Figure \ref{fig:main_results} compares performance of multi-agent frameworks across various single-agent and multi-agent baselines (varies across benchmarks - some single-agent baselines for GPQA are obtained from \url{https://klu.ai/glossary/gpqa-eval}). From the results, it is evident that the multi-agent frameworks are consistently outperforming the baselines.

\paragraph{Performance on NATURAL PLAN}
From Figure \ref{subfig:natural_plan}, \plangen{} (Best of $\mathcal{N}$) achieves the highest EM scores across all tasks: $60.70$ (Calendar), $43.80$ (Meeting), and $41.63$ (Trip). In calendar scheduling, all four frameworks surpass the strongest baseline (Multi-Agent Baseline) by $\sim10\%$. For meeting and trip planning, all except ToT outperform the best baseline (Gemini-1.5-Pro) by $\sim6\%$ and $\sim7\%$, respectively. \plangen{} (Mixture of Algo.) achieves the second-highest performance in meeting and trip planning while remains competitive in calendar scheduling. These results demonstrate the effectiveness of our frameworks in handling diverse natural language planning tasks and establishing SOTA for all three categories of NATURAL PLAN.

\paragraph{Performance on OlympiadBench}
From Figure \ref{subfig:olympiad}, \plangen{} (Mixture of Algo.) achieves the highest accuracy in the MATH category ($55.94\%$), outperforming the strongest Multi-Agent Baseline ($50.68\%$) by $\sim5\%$. Notably, the superior performance of the \plangen{} (Mixture of Algo.) in MATH highlights its effectiveness in complex mathematical reasoning, setting a SOTA for the MATH. In the PHY category, all multi-agent frameworks surpass Gemini-1.5-Flash (strongest baseline), with \plangen{} (Best of $\mathcal{N}$) achieving the highest accuracy ($31.78\%$), setting a SOTA for the PHY.

\paragraph{Performance on GPQA}
From Figure \ref{subfig:gpqa}, the \plangen{} (Mixture of Algo.) achieves the highest accuracy ($59.6\%$). The individual inference-time algorithms achieve a lower performance, indicating the usefulness of selection. All proposed frameworks outperform Gemini-1.5-Pro ($46.2\%$), GPT models ($\sim48\%$), and Claude-3-Opus ($50.4\%$) by a large margin. While Claude-3.5-Sonnet, and Multi-Agent Baseline perform competitively ($\sim59\%$) compared to \plangen{} (Mixture of Algo.).

\paragraph{Performance on DocFinQA}
From Figure \ref{subfig:docfinqa}, our frameworks significantly improve performance on DocFinQA, with \plangen{} (Best of $\mathcal{N}$) achieving the highest accuracy ($31.16\%$) and F1-Score ($29.45\%$), setting SOTA for the task. All our frameworks outperform the Gemini-1.5-Pro (strongest baseline) by a large margin ($\sim7\%$). These results highlight the effectiveness of multi-agent frameworks in financial document understanding, and performing reasoning over them.

\begin{wrapfigure}{r}{0.5\textwidth}
    \centering
    \includegraphics[width=\linewidth]{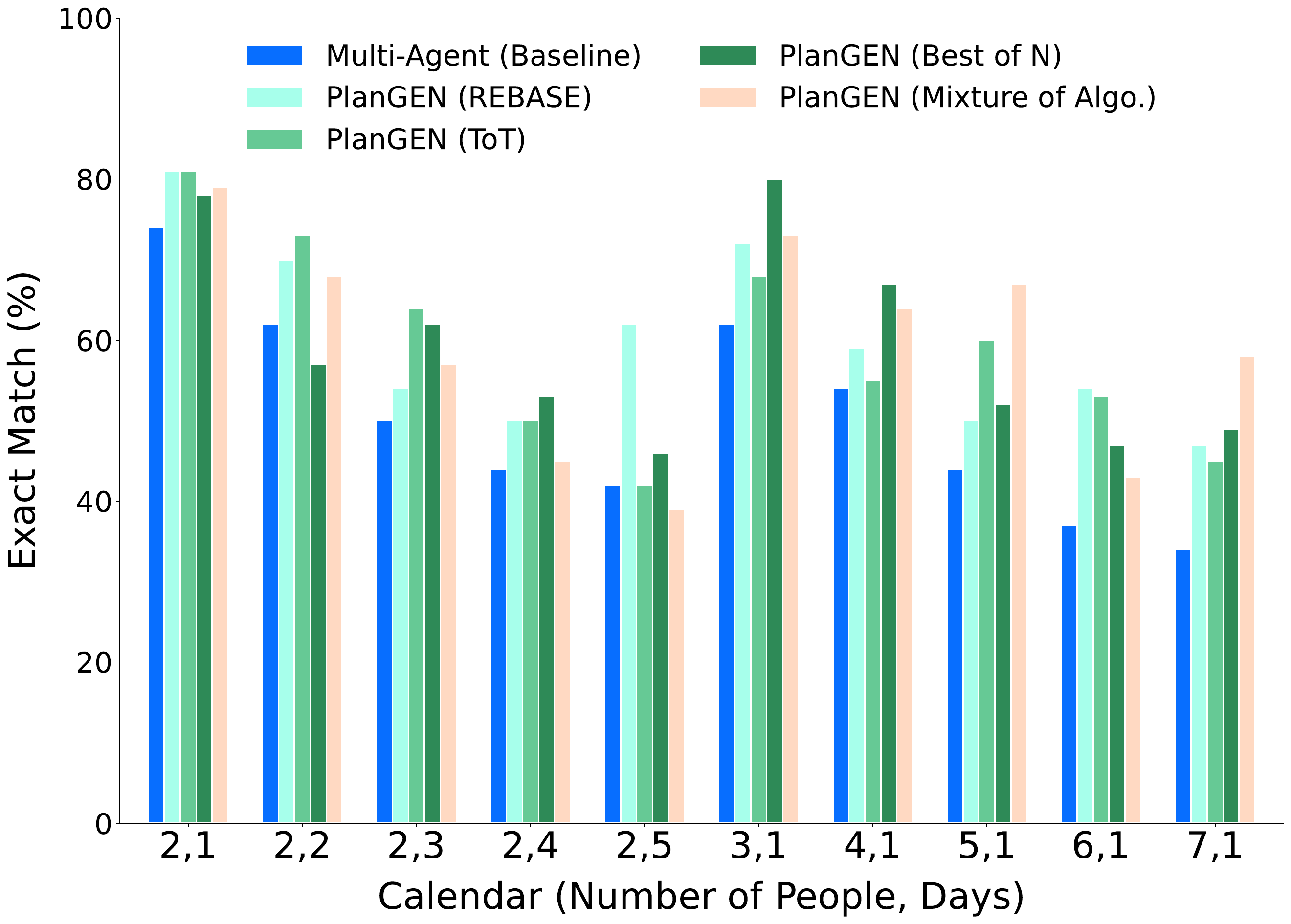}
    \caption{Performance comparison of inference-time algorithms across different complexity levels for calendar scheduling from NATURAL PLAN.}
    \label{fig:np_cal_analysis}
\end{wrapfigure}

\paragraph{Performance of our frameworks w.r.t. different complexity}
As shown in Figure \ref{fig:np_cal_analysis}, we conduct a case study on calendar scheduling task from NATURAL PLAN to analyze the impact of varying complexity levels on the performance of different frameworks. For the calendar scheduling, we observe that \plangen{} (ToT) performs best for simple problems, while \plangen{} (Best of $\mathcal{N}$) is more effective for intermediate problems. As complexity increases, a \plangen{} (Mixture of Algo.) proves to be the most effective approach. We further conduct a similar analysis for meeting and trip planning from NATURAL PLAN presented in App. \ref{app:analysis}.


\paragraph{Main Findings} 
Compared to single-agent systems, multi-agent frameworks consistently outperform in generating optimized planning trajectories (Figure \ref{fig:main_results}). Furthermore, Multi-Agent (Baseline) is not always the strongest benchmark, as self-correction can introduce challenges as shown in \citet{huang2024large}. Thus, different agents within the system require distinct handling strategies similar to our \plangen{}. Additionally, even in multi-agent frameworks for \plangen{}, relying on a single inference-time algorithm proves insufficient for more complex problems (Figure \ref{fig:np_cal_analysis}). A \plangen{} (Mixture of Algo.) approach offers substantial advantages for solving complex planning problems, highlighting the importance of algorithm selection based on instance-specific complexity (Figure \ref{fig:teaser}). Given that our frameworks are multi-agent, we provide further discussion on $\#$ of LLM calls \textit{vs.} their performance in subsequent section.

\section{Analysis and Discussion}
\label{sec:discussion}

Here, we discuss detailed analysis over importance of our agents and model-agnostic nature of our frameworks. Additionally, we also present more analysis on results in App. \ref{app:analysis}.

\paragraph{Importance of Verification Agent}

\begin{wrapfigure}{r}{0.5\textwidth}
    \centering
    \includegraphics[width=\linewidth]{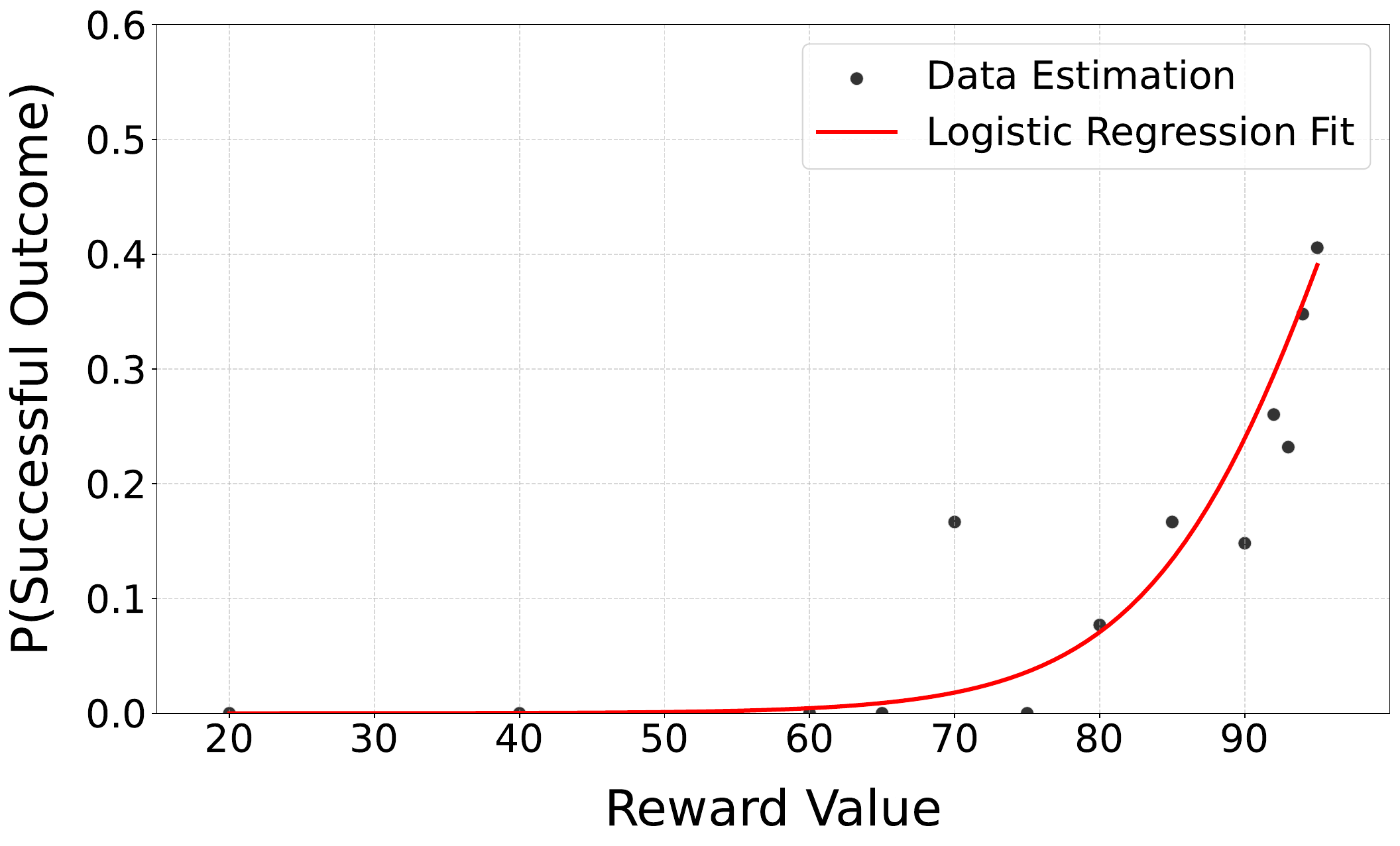}
    \caption{Logistic regression plot showing verification agent's positive performance impact. P(Successful Outcome) $=$ probability of prediction being correct.}
    \label{fig:imp_verification}
\end{wrapfigure}Figure \ref{fig:imp_verification} demonstrates the verification agent's crucial role in \plangen{} by showing a strong correlation between assigned reward values and prediction correctness (1 for correct, 0 for incorrect).  The plotted points represent the average correctness rate for data buckets of varying reward values, each bucket containing hundreds of samples. A logistic regression model trained on DocFinQA and GPQA data ($\sim1100$ total samples) reveals a sigmoidal trend: higher rewards correlate with increased success probability, highlighting the agent's effectiveness.  This reinforces the importance of constraint-guided verification for improving inference-time algorithms (see App. \ref{app:analysis} for further details).


\paragraph{Importance of Selection Agent}

\begin{wrapfigure}{r}{0.5\textwidth}
    \centering
    \includegraphics[width=\linewidth]{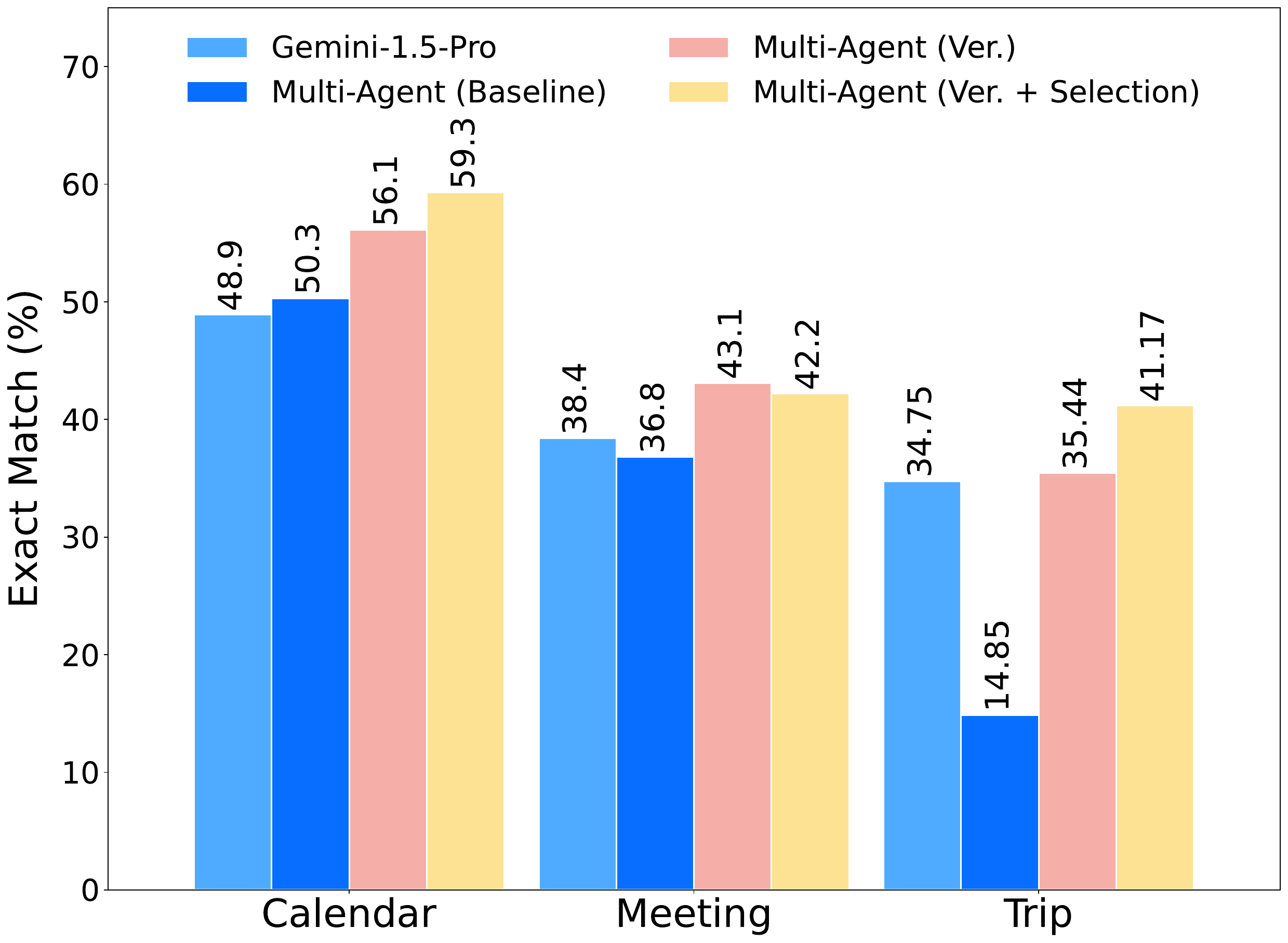}
    \caption{Case study on NATURAL PLAN, showing the impact of selection agent. Ver.: Verification.}
    \label{fig:imp_selection}
\end{wrapfigure}

Figure \ref{fig:imp_selection} illustrates the importance of the selection agent by comparing the performance on the NATURAL PLAN. Here, Multi-Agent (Ver.) includes only the verification agent, while Multi-Agent (Ver. + Selection) further includes a selection agent. The results highlight the progressive impact of these components. 

For example, in calendar scheduling, Multi-Agent (Ver.) improves performance to 56.1 EM compared to Multi-Agent (Baseline). However, Multi-Agent (Ver. + Selection) achieves 59.3 EM, demonstrating the additional benefit of algorithm selection. A similar trend is observed in trip planning where Multi-Agent (Ver. + Selection) outperforms Multi-Agent (Ver.) (41.17 EM vs. 35.44 EM) and the Multi-Agent (Baseline). For meeting planning, Multi-Agent (Ver.) achieves 43.1 EM compared to 36.8 EM of Multi-Agent (Baseline), whereas, Multi-Agent (Ver. + Selection) achieves competitive performance. Together, verification and selection agents drive significant improvements over single-agent and multi-agent baselines.

\begin{table}
\resizebox{\linewidth}{!}{
\begin{tabular}{c|c|c|cc||c|c|c|cc||c|c}
\toprule
                                                                                &                                 &                                                                                              & \multicolumn{2}{c||}{\textbf{OlympiadBench}}                         &                                                                                 &                                 &                                                                                              & \multicolumn{2}{c||}{\textbf{OlympiadBench}}                         &                                                                             &                                 \\ \cmidrule{4-5} \cmidrule{9-10}
\multirow{-2}{*}{\textbf{Methods}}                                              & \multirow{-2}{*}{\textbf{GPQA}} & \multirow{-2}{*}{\textbf{\begin{tabular}[c]{@{}c@{}}NATURAL PLAN\\ (Calendar)\end{tabular}}} & \multicolumn{1}{c|}{\textbf{MATH}}                 & \textbf{PHY}   & \multirow{-2}{*}{\textbf{Methods}}                                              & \multirow{-2}{*}{\textbf{GPQA}} & \multirow{-2}{*}{\textbf{\begin{tabular}[c]{@{}c@{}}NATURAL PLAN\\ (Calendar)\end{tabular}}} & \multicolumn{1}{c|}{\textbf{MATH}}                 & \textbf{PHY}   & \multirow{-2}{*}{\textbf{Methods}}                                          & \multirow{-2}{*}{\textbf{GPQA}} \\ \midrule
\rowcolor[HTML]{EFEFEF} 
Gemini-1.5-Pro                                                                  & 46.20                           & 48.90                                                                                        & \multicolumn{1}{c|}{\cellcolor[HTML]{EFEFEF}32.63} & 28.35          & Gemini-2.0-Flash                                                                  & 60.10                           & 61.10                                                                                        & \multicolumn{1}{c|}{\cellcolor[HTML]{EFEFEF}51.13} & 27.54          & GPT-4o                                                                      & 47.98                           \\ \midrule
\begin{tabular}[c]{@{}c@{}}PlanGEN (BoN)\\ Gemini-1.5-Pro\end{tabular}              & 56.60                           & \textbf{60.70}                                                                               & \multicolumn{1}{c|}{53.85}                         & \textbf{31.78} & \begin{tabular}[c]{@{}c@{}}PlanGEN (BoN)\\ Gemini-2.0-Flash\end{tabular}              & 56.83                           & \textbf{68.90}                                                                               & \multicolumn{1}{c|}{59.90}                         & 35.60          & \begin{tabular}[c]{@{}c@{}}PlanGEN (BoN)\\ GPT-4o\end{tabular}              & 40.40                           \\ \midrule
\begin{tabular}[c]{@{}c@{}}PlanGEN (ToT)\\ Gemini-1.5-Pro\end{tabular}              & 56.60                           & 59.10                                                                                        & \multicolumn{1}{c|}{54.45}                         & 29.37          & \begin{tabular}[c]{@{}c@{}}PlanGEN (ToT)\\ Gemini-2.0-Flash\end{tabular}              & 59.18                           & 62.30                                                                                        & \multicolumn{1}{c|}{60.30}                         & 35.70          & \begin{tabular}[c]{@{}c@{}}PlanGEN (ToT)\\ GPT-4o\end{tabular}              & 46.70                           \\ \midrule
\begin{tabular}[c]{@{}c@{}}PlanGEN (REBASE)\\ Gemini-1.5-Pro\end{tabular}           & 57.10                           & 59.90                                                                                        & \multicolumn{1}{c|}{54.90}                         & 31.36          & \begin{tabular}[c]{@{}c@{}}PlanGEN (REBASE)\\ Gemini-2.0-Flash\end{tabular}           & \textbf{64.14}                  & 61.50                                                                                        & \multicolumn{1}{c|}{60.98}                         & 36.02          & \begin{tabular}[c]{@{}c@{}}PlanGEN (REBASE)\\ GPT-4o\end{tabular}           & 41.40                           \\ \midrule
\begin{tabular}[c]{@{}c@{}}PlanGEN (Mixture of Algo.)\\ Gemini-1.5-Pro\end{tabular} & \textbf{59.60}                  & 59.30                                                                                        & \multicolumn{1}{c|}{\textbf{55.94}}                & 31.28          & \begin{tabular}[c]{@{}c@{}}PlanGEN (Mixture of Algo.)\\ Gemini-2.0-Flash\end{tabular} & 63.64                           & 66.55                                                                                        & \multicolumn{1}{c|}{\textbf{64.10}}                & \textbf{37.29} & \begin{tabular}[c]{@{}c@{}}PlanGEN (Mixture of Algo.)\\ GPT-4o\end{tabular} & \textbf{49.40}                  \\ \bottomrule
\end{tabular}
}
\caption{Performance comparison for model-agnostic nature of \plangen{}.  We utilize Gemini-1.5-Pro, Gemini-2.0-Flash, and GPT-4o as baseline and underlying model in \plangen{} frameworks. Comparing methods that use the same base and underlying model for a fair assessment}
\label{tab:model_agnostic}
\end{table}

\paragraph{Model-Agnostic Nature}


The results from Table \ref{tab:model_agnostic} demonstrate the model-agnostic nature of our proposed multi-agent frameworks. While the primary experiments were conducted using Gemini-1.5-Pro, the framework's effectiveness holds across different underlying models, such as Gemini-2.0-Flash and GPT-4o. For instance, in the NATURAL PLAN (calendar scheduling), the \plangen{} (Best of $\mathcal{N}$) framework achieves a significant improvement, reaching 68.90 EM, outperforming Gemini-2.0-Flash (61.10 EM). Similarly, in OlympiadBench, the \plangen{} (Mixture of Algo.) achieves the highest scores in MATH (64.10) and PHY (37.29), surpassing Gemini-2.0-Flash (52.13 MATH, 27.54 PHY). Note that, the Mixture of Algo. outperforms other three frameworks, showing effectiveness of selection agent. On GPQA, Mixture of Algo. (49.40) and \plangen{} (REBASE) (64.14) outperform GPT-4o (47.98) and Gemini-2.0-Flash (60.10), respectively. These results highlight that regardless of the underlying model, our frameworks consistently enhance performance by leveraging multi-agent collaboration, reinforcing their flexibility and robustness across models. 





\paragraph{Discussion on LLM calls \textit{vs.} Performance (\%)}

\begin{wrapfigure}{r}{0.5\textwidth}
    \centering
    \vspace{-10mm}
    \includegraphics[width=\linewidth]{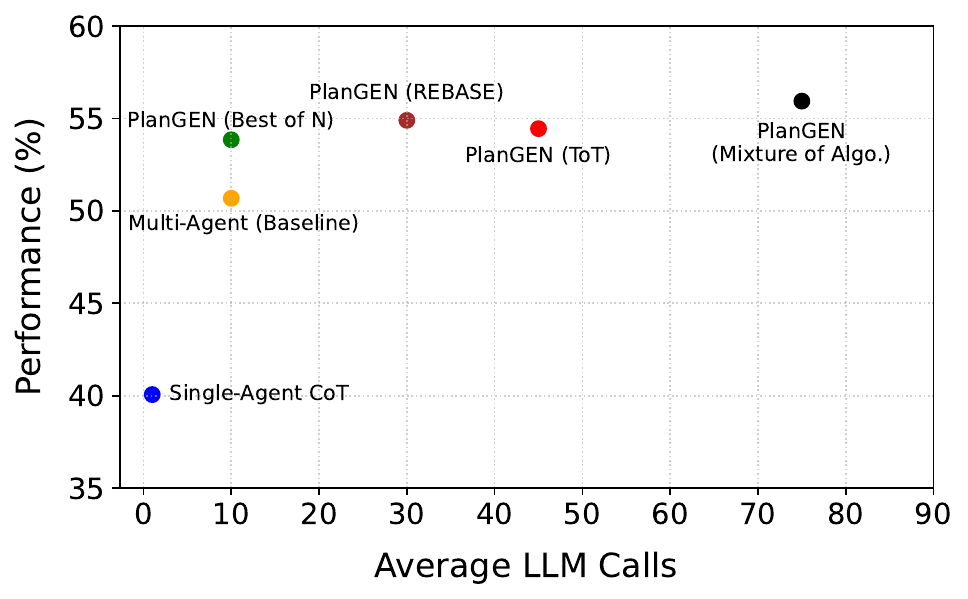}
    \caption{Comparison of baselines and our frameworks, showing the trade-off between LLM calls and performance (\%) for OlympiadBench (MATH).}
    \label{fig:llm_calls_vs_performance}
\end{wrapfigure}Figure \ref{fig:llm_calls_vs_performance} shows the relationship between the number of LLM calls and task performance across baselines (single-agent and multi-agent) and proposed frameworks, using OlympiadBench (MATH category). The single-agent system, zero-shot CoT, requires only one LLM call. The multi-agent baseline requires the same number of calls as \plangen{} (Best of $\mathcal{N}$), but our framework outperforms the multi-agent baseline. For \plangen{} (ToT) and \plangen{} (REBASE), we focus on LLM calls during the tree expansion phase. \plangen{} (ToT) involves dynamic exploration, where each explored path requires three LLM calls: step generation, reward evaluation, and completion verification. The total cost is the per-path cost multiplied by the number of paths explored, constrained by either the number of steps generated for each problem or a predefined iteration budget (i.e., 20). For \plangen{} (REBASE), the number of LLM calls depends on the search width (i.e., 10). Each solution path expansion involves three calls: step generation, quality evaluation, and completion verification, thus, giving maximum 30 LLM calls for single problem. For \plangen{} (Mixture of Algo.), we estimate the average LLM calls by summing the estimated calls for each selected algorithm per problem, then dividing by the total number of problems. As shown in Figure \ref{fig:llm_calls_vs_performance}, the single-agent system exhibits the lowest performance despite requiring just one LLM call. Multi-agent approaches show improved performance, with \plangen{} (ToT) and \plangen{} (REBASE) balancing LLM call efficiency and accuracy. The \plangen{} (Mixture of Algo.) method achieves the highest performance, suggesting that combining diverse planning strategies enhances efficiency.

\section{Conclusions}
\label{sec:conclusions}

In this work, we proposed \plangen{}, an easily scalable multi-agent approach incorporating three key components: constraint, verification, and selection agents. We leveraged these agents to improve the verification process of existing inference algorithms and proposed three frameworks: Multi-Agent Best of $\mathcal{N}$, ToT, and REBASE. Further, we introduced a Mixture of Algorithms, an iterative framework that integrates the selection agent (Figure \ref{fig:teaser}) to dynamically choose the best algorithm. We evaluated our frameworks on NATURAL PLAN, OlympiadBench, GPQA, and DocFinQA. Experimental results demonstrate that \plangen{} outperforms strong baselines, achieving SOTA results across datasets. Furthermore, our findings suggest that the proposed frameworks are scalable and generalizable to different LLMs, improving their natural language planning ability.

\section*{Limitations}

Despite the strong performance of our frameworks, an area of improvement is the reliance on predefined heuristics for selecting inference-time algorithms, which may not always generalize optimally across all tasks and domains. Additionally, while our frameworks demonstrate strong performance, their computational overhead could be further optimized for efficiency in real-world applications. We believe that our frameworks can be useful in further boosting the planning and reasoning capabilities of existing models such as o1 and Gemini-thinking. In addition, the use of reinforcement learning or meta-learning techniques to dynamically adapt agent strategies based on task complexity could be an interesting area to explore. Moreover, broadening the scope to multi-modal and multi-lingual reasoning would significantly expand the applicability of our approach, and exploring the use of generated planning trajectories for model training offers valuable direction.

\section*{Ethics Statement}

The use of proprietary LLMs such as GPT-4, Gemini, and Claude-3 in this study adheres to their policies of usage. We have used AI assistants (Grammarly and Gemini) to address the grammatical errors and rephrase the sentences.


\bibliographystyle{abbrvnat}
\nobibliography*
\bibliography{main}


\clearpage

\appendix

\section{Related Works}
\label{sec:related_works}

\paragraph{LLM Agents for Planning}

Agent-based frameworks for planning have gained interest, focusing on enhancing how LLMs decompose tasks and refine their outputs. The Sibyl framework \citep{wang2024sibyl} effectively decomposes tasks into smaller subtasks, assigning each to specialized agents that iteratively collaborate until a solution is reached. OS-Copilot \citep{wu2024copilot} introduces a generalist computer agent that employs self-improvement through modularization and feedback loops. Another approach is KnowAgent \citep{zhu2024knowagent}, which integrates knowledge-augmented planning to enhance the decision-making capabilities of LLM agents. Similarly, Tool-Planner \citep{liu2024tool} proposed grouping tools based on similar functionalities into toolkits, allowing LLMs to select the best tool for a given task. Many agent-based works focusing on planning have been developed \citep{chen2024reprompt, xie2024human, wang2024promptagent}. Despite the progress, these methods generally (i) focus on domain-specific tasks or limited benchmarks, reducing generalizability, and (ii) lack or under-explore mechanisms for verifying and refining plans iteratively. While some works explore natural language planning \citep{bohnet2024exploring, lee2025evolving}, they either single-agent frameworks or evaluate proposed framework on domain-specific benchmarks.  


\paragraph{Inference-time Algorithms}

Inference-time algorithms have recently shown a significant improvement in LLMs performance during inference. For instance, Best of $\mathcal{N}$ sampling \citep{brown2024large} selects the most promising output from multiple generations performed using temperature sampling, while Tree-of-Thought (ToT) \citep{yao2024tree} models reasoning as an iterative tree search. REBASE \citep{wu2024empirical} optimizes search-space pruning using reward balancing. One very popular approach is Monte Carlo Tree Search (MCTS) \citep{zhang2024accessing} which iteratively explores solution paths during inference. Applied to models such as LLaMa-3-8B, it enables self-refinement by revisiting and improving initial solutions. Test-time optimization \citep{snell2024scaling}, focuses on dynamically adjusting computational resources during inference \citep{wu2024empirical}. Furthermore, \citet{wang2025planning} uses the inference time algorithms to improve LLMs planning capabilities to solve code synthesis problems. In inference-time algorithms, verification is the key component. In contrast to these past works, here, we enhance performance of inference-time algorithms utilizing constraint-guided verification, and multi-agent collaboration for natural language planning, and its applications in downstream complex reasoning tasks.

\section{Further Details on LLM Agents}
\label{app:llm_agents}

In this section, we provide additional details about each specialized agent in \plangen{}. We present the prompts used for each agent, highlighting their roles in the framework. The prompt for the constraint agent includes task-specific parameters that can be adjusted to extract relevant constraints for different tasks. In contrast, the prompts for the verification agent and selection agent are entirely task-agnostic, ensuring generalizability and adaptability across various problem domains.

\paragraph{Prompts for Constraint Agent}
The constraint agent is responsible for extracting problem-specific constraints that guide the planning process. To enable systematic extraction of constraints, we design a task-specific prompt for the constraint agent:


\begin{tcolorbox}[boxrule=0pt, frame hidden, title=Prompt, breakable, sharp corners, borderline west={0pt}{0pt}{black!50}, title style={
        colback=black!50, 
        colframe=black!50, 
        coltitle=black 
    }]
You are an expert in understanding an input problem and generating set of constraints. Analyze the input problem and extract all relevant instance-specific constraints and contextual details necessary for accurate and feasible planning. 
\\
\\
(\hl{Optional}) These constraints may include:
\\

<You may provide any specific type of constraints>
\\

<You may provide any formatting instruction>
\\

\textbf{Input Problem:} <problem statement>
\end{tcolorbox}

Based on the above prompts, we define the types of constraints used in the NATURAL PLAN benchmark for different planning tasks: calendar scheduling, meeting planning, and trip planning. For DocFinQA, we provide a set of formatting instructions to ensure structured constraint generation. For GPQA and OlympiadBench, the constraint extraction follows the general prompt outlined above.

\paragraph{Prompts for Verification Agent}

The prompt for the verification agent is designed to be task-agnostic, meaning it can be applied across different problem domains without modification. By enforcing strict evaluation criteria, this agent enhances the reliability of \plangen{}, making it robust for various planning and reasoning tasks. In this prompt, list of constraints are generated using constraint agent. Notably, the list of constraints used in the verification prompt is dynamically generated by the constraint agent. This ensures that the verification process is based on instance-specific constraints rather than relying on predefined, static rules.

\begin{tcolorbox}[boxrule=0pt, frame hidden, title=Prompt, breakable, sharp corners, borderline west={2pt}{0pt}{black!50}, title style={
        colback=black!50, 
        colframe=black!50, 
        coltitle=black 
    }]

Provide a reward score between -100 and 100 for the quality of the provided plan steps, using strict evaluation standards. Ensure the reward reflects how effectively the plan contributes to progressing toward the correct solution.

\textbf{Problem Statement:}

\{problem\}

\textbf{Plan:}

\{plan\}

\textbf{Consider the following constraints while evaluating:}

- [Constraint 1]

- [Constraint 2]

- [Constraint 3]

\textbf{Provide feedback in the following format:}

[Step-by-step reasoning for the reward score]

\textbf{Score:} [Strictly provide an integer reward score between -100 and 100]
\end{tcolorbox}

\paragraph{Prompts for Selection Agent}

The prompt for the Selection Agent is task-agnostic, allowing it to be applied across various domains without modification. It processes feedback from the verification agent and contextual information from the problem statement to assign suitability scores to different inference-time algorithms.

\begin{tcolorbox}[boxrule=0pt, frame hidden, title=Prompt, breakable, sharp corners, borderline west={0pt}{0pt}{black!50}, title style={
        colback=black!50, 
        colframe=black!50, 
        coltitle=black 
    }]

Analyze the following planning problem and explain your reasoning for assigning priority scores to the algorithms based on their suitability. Scores should be between 0 and 1, where 1 represents the most suitable algorithm for the given problem.

\textbf{Problem Statement:} <problem statement>

\textbf{Requirements:} <feedback>

\textbf{Context:} <context if context else `None provided'>

Start by providing a brief reasoning for each algorithm's suitability based on problem complexity. Then, \textbf{ONLY output your response strictly as a list} with the exact format below:

\textbf{Reasoning:}
\begin{itemize}
    \item \textbf{Best of N:} [Explain why this algorithm is or isn’t suitable]
    \item \textbf{Rebase:} [Explain why this algorithm is or isn’t suitable]
    \item \textbf{ToT:} [Explain why this algorithm is or isn’t suitable]
\end{itemize}

\textbf{Scores:}
\begin{equation*}
\begin{aligned}
&[\text{("Best of N", float)}, \\
&\text{("Rebase", float)}, \\
&\text{("ToT", float)}]
\end{aligned}
\end{equation*}

\end{tcolorbox}

\paragraph{Algorithm for Selection using UCB}

The algorithm (Algorithm \ref{algo:selection}) presented is a modified UCB selection strategy that incorporates additional factors for exploration, diversity, and recovery. It initializes each algorithm with basic statistics like reward ($R(a)$), count of trials ($C(a)$), and recovery score ($Rec(a)$). The algorithm computes a normalized reward $\bar{R}_{\text{norm}}(a)$ for each option, balancing the reward with exploration ($E(a)$), which encourages trying less-used algorithms. A diversity bonus $D(a)$ penalizes overused algorithms, while a recovery bonus $RecB(a)$ rewards algorithms that perform well after prior failures. LLM-guided priors ($LLM\_prior$) are used to influence the selection process based on prior knowledge. The final selection is made by maximizing the UCB score, which combines these factors to balance exploitation and exploration.

\paragraph{Ablation Study on UCB Modifications}
\begin{wrapfigure}{r}{0.5\textwidth}
    \centering
    \vspace{-9mm}
    \includegraphics[width=\linewidth]{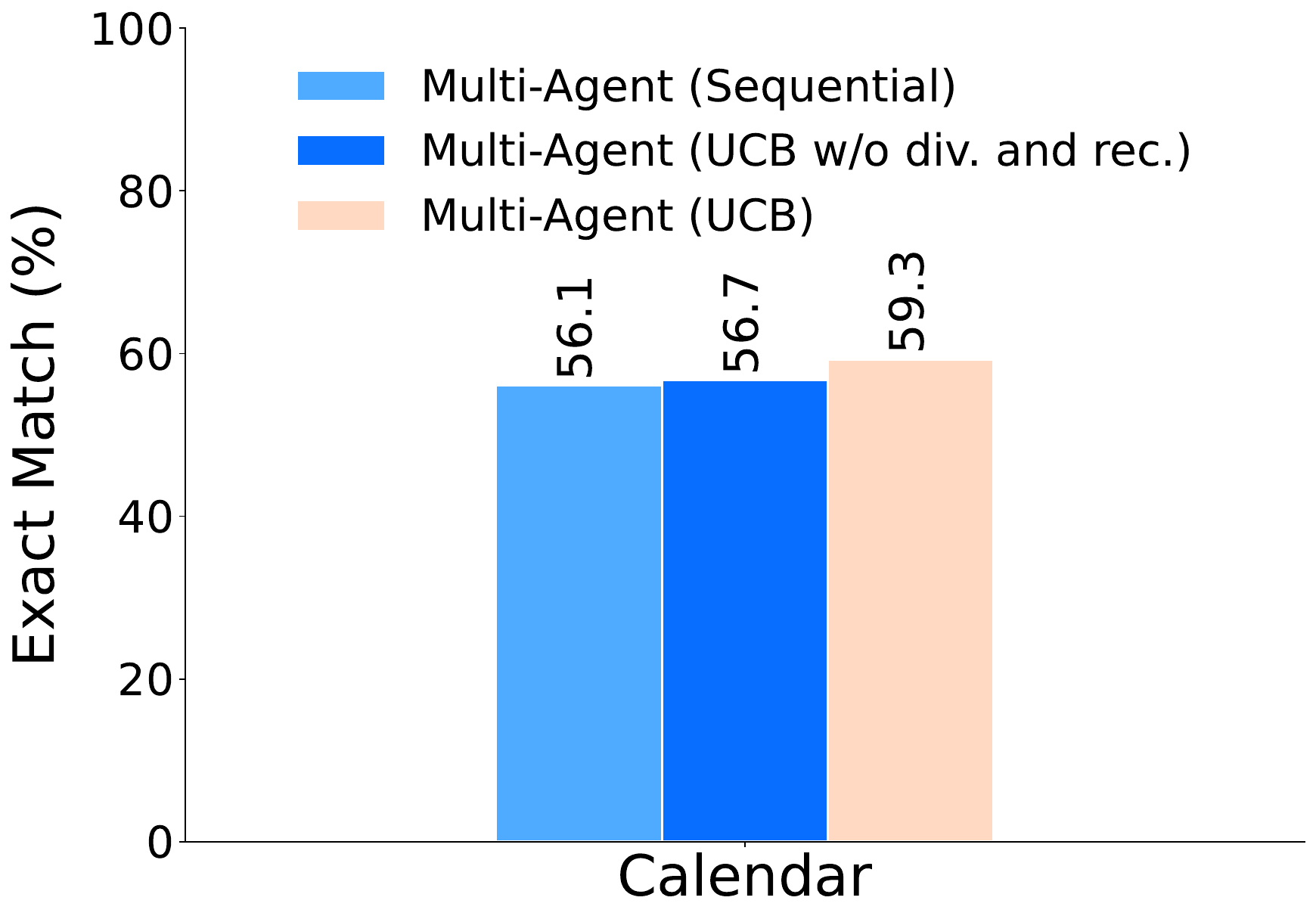}
    \caption{Ablation Study of UCB Modifications on Selection Agent and its impact on Multi-Agent Mixture of Algorithms framework. div.: diversity bonus, rec: recovery term.}
    \label{fig:ucb_analysis}
\end{wrapfigure}
To design our selection agent, we conducted an ablation study evaluating modifications to the UCB formula, shown in Figure \ref{fig:ucb_analysis}.  Initially, we replaced the selection agent with a simple sequential strategy, termed ``Multi-Agent (Sequential)'', where algorithms execute in sequence, and the verification agent selects the highest-scoring plan.  Next, we implemented a UCB selection agent, but excluded the `diversity bonus' and `recovery term' introduced in our proposed formulation in the main paper, denoted as ``Multi-Agent (UCB w/o div. and rec.)''. Finally, we implemented the complete selection agent incorporating our proposed UCB, labeled ``Multi-Agent (UCB)''.  As shown in Figure \ref{fig:ucb_analysis}, the inclusion of the diversity bonus and recovery terms in the UCB formula ("Multi-Agent (UCB)") resulted in $\sim3.5\%$ performance gain compared to the UCB variant without these terms, further enhancing overall results. Note that the LLM-guided priors are still the part of Multi-Agent (UCB w/o div. and rec.) and Multi-Agent (UCB).

\section{Prompts for Proposed Frameworks}
\label{app:frameworks}

We provide further details in this section regarding the prompts used for \plangen{} (ToT) and \plangen{} (REBASE), as well as the specific algorithms used to execute these inference-time methods.



\paragraph{Prompts used for ToT and REBASE}

\plangen{} (ToT) and \plangen{} (REBASE) employ three prompt types: (1) step prompt, (2) step reward prompt, and (3) completion prompt. Step prompt guide the model to generate subsequent steps based on the problem statement and previously generated steps. Step reward prompt evaluate each intermediate step against the problem statement and constraints, similar to the prompts used by a verification agent. Completion prompt check for a complete solution after each step. If a solution is found, exploration terminates; otherwise, the process continues until a solution is reached.

\begin{tcolorbox}[boxrule=0pt, frame hidden, title=Step Prompt, breakable, sharp corners, borderline west={0pt}{0pt}{black!50}, title style={
        colback=black!50, 
        colframe=black!50, 
        coltitle=black 
    }]

You are an expert assistant for generating step-by-step plan to solve a given question using specified tools. Given the problem and any intermediate steps, output only the next step in the plan. Ensure that the next action helps in moving toward the correct plan to solve the given question. Do not provide the full plan. Keep responses concise, focusing solely on the immediate next step that is most effective in progressing toward the correct plan.
\\
\\
<problem>

\{Add a problem statement here\}

</problem>
\\
\\
<intermediate\_step>

\{Append previously generated steps\}

</intermediate\_step>
\end{tcolorbox}

\begin{tcolorbox}[boxrule=0pt, frame hidden, title=Completion Prompt, breakable, sharp corners, borderline west={0pt}{0pt}{black!50}, title style={
        colback=black!50, 
        colframe=black!50, 
        coltitle=black 
    }]

You are an assistant tasked with verifying if the final, complete plan to solve the given question has been achieved within the intermediate steps. Output only `1' if the intermediate steps contain the full solution needed to solve the question. If the full plan has not yet been reached, output only `0'. Provide no additional commentary—return exclusively `1' or `0'.
\\
\\
<problem>

\{Add a problem statement here\}

</problem>
\\
\\
<intermediate\_step>

\{Append previously generated steps\}

</intermediate\_step>
\end{tcolorbox}

\section{Details on Benchmarks and Experiments}
\label{app:experiments}

\paragraph{Statistics of Benchmarks}

For evaluation, we utilize evaluation sets of all four benchmarks. For NATURAL PLAN, we employed the provided evaluation sets, consisting of 1000 instances each for Calendar Scheduling and Meeting Planning, and 1600 instances for Trip Planning.  The GPQA evaluation was conducted using the Diamond set, which comprises 198 highly challenging instances.  From OlympiadBench, we selected the text-only problems, excluding those requiring a theorem prover, resulting in 674 instances for the MATH category and 236 for the PHY category. We also used 922 instances from the DocFinQA evaluation set.

\paragraph{Models}
Our primary evaluations use Gemini-1.5-Pro for all the experiments.  We also present a case study with Gemini-2.0-Flash and GPT-4o to showcase the model-agnostic nature of \plangen{}.

\paragraph{Metrics}
We use task-specific metrics for all evaluations. Specifically, we use Exact Match (EM) for NATURAL PLAN similar to \citet{zheng2024natural}, micro-average accuracy for OlympiadBench similar to \citet{he-etal-2024-olympiadbench}, and accuracy for GPQA and DocFinQA (along with F1-Score for DocFinQA).

\paragraph{Feedback prompt for Multi-Agent Baseline} In the multi-agent baseline, we employ a feedback prompt to iteratively generate improved and refined outputs. The prompt is provided below:

\begin{tcolorbox}[boxrule=0pt, frame hidden, title=Feedback Prompt, breakable, sharp corners, borderline west={0pt}{0pt}{black!50}, title style={
        colback=black!50, 
        colframe=black!50, 
        coltitle=black 
    }]

Analyze the following planning problem and explain your reasoning for assigning priority scores You are an intelligent assistant capable of self-reflection and refinement. I will provide you with your last response, and your task is to improve it, if necessary.

Here is your previous response:

\{previous\_response\}

Analyze and refine your response step-by-step:

1. Reflect on your reasoning process. Where might it be unclear or incorrect? Improve it.

2. Revise the explanation to address any identified issues and make it more logical and comprehensive.

3. Ensure the final answer is correct, supported by clear reasoning.
\end{tcolorbox}

\paragraph{Hyper-parameters for Experiments}
To ensure deterministic behavior, we set the temperature of all models to 0 for each agent.  For the inference-time algorithms, we used the following settings: \plangen{} (Best of $\mathcal{N}$) with five samples at a temperature of 0.7; Tree of Thoughts (ToT) with three children per root node, generated at a temperature of 0.7; and REBASE, initialized with width $10$ at temperature of 0.7, decremented by 1 after each call to expand.

\section{Additional Analysis}
\label{app:analysis}

\paragraph{Importance of Verification Agent}
The kernel density estimation (KDE) plot visualizes the distribution of reward values assigned to two distinct outcomes: ``Success'' (green) and ``Failure'' (red).  The plot reveals a clear separation between the reward distributions, with ``Success'' outcomes strongly associated with high reward values (around 80-100) and ``Failure'' outcomes primarily associated with low reward values (around 20-40). The sharply peaked green curve suggests consistent and high rewards for successful outcomes, while the broader red curve reflects more variability in rewards assigned to failures.  However, a small bump in the red curve at high reward values (around 80-90) suggests a few instances where failures received unexpectedly high rewards, warranting further investigation. 
\begin{wrapfigure}{r}{0.5\textwidth}
    \centering
    \vspace{2mm}
    \includegraphics[width=\linewidth]{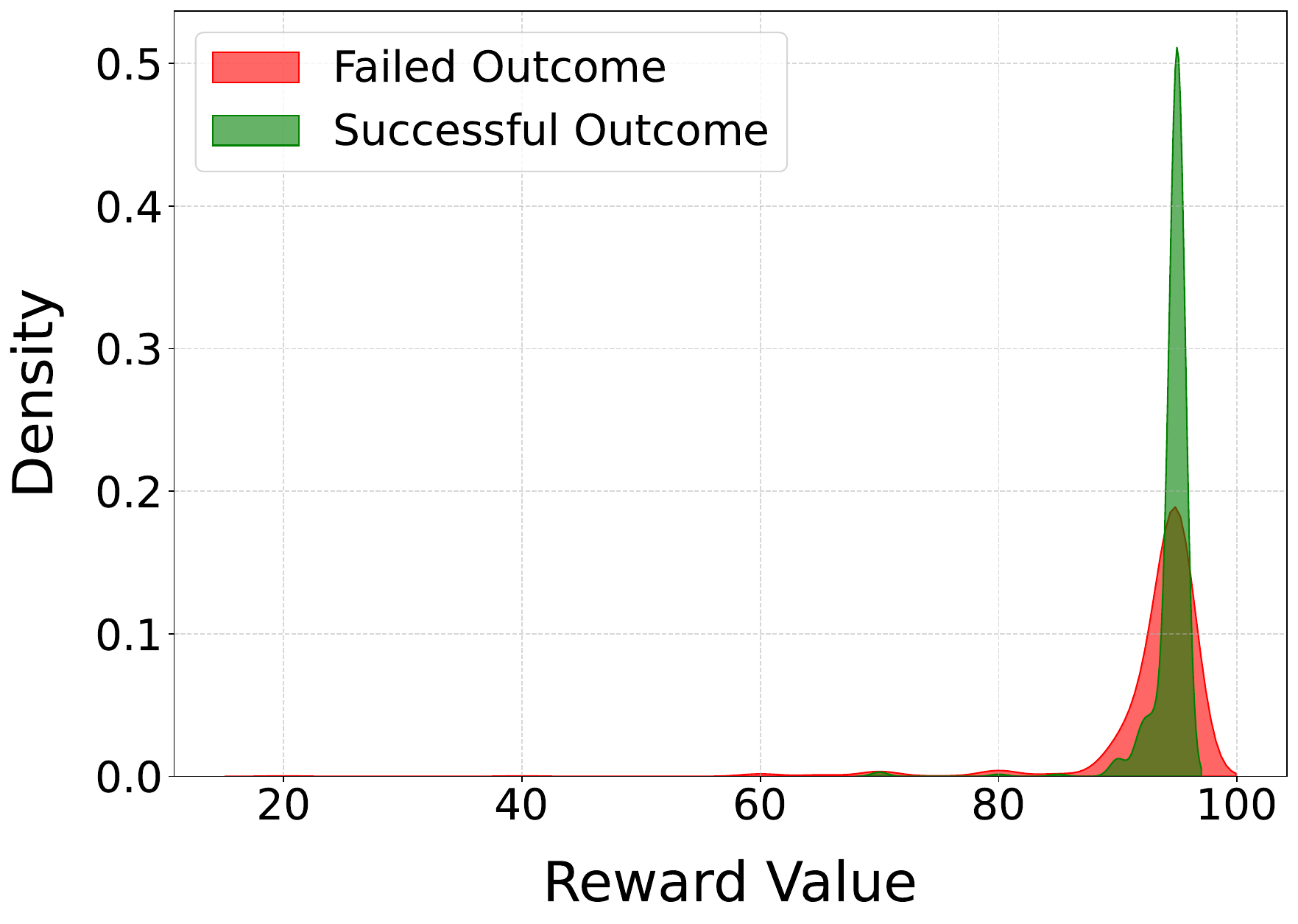}
    \caption{KDE plot illustrating relationship between reward value and outcome (success/failure).}
    \label{fig:kde_plot}
\end{wrapfigure}This observation is further supported by a statistically significant difference between the reward distributions, a Mann-Whitney U test ($U = 116128.0$, $p < 0.0001$). The low p-value (3.42e-09) provides evidence that the difference in reward distributions is statistically significant.

\paragraph{Performance of our frameworks w.r.t. different complexity}

\begin{figure}
    \centering
    \includegraphics[width=\textwidth]{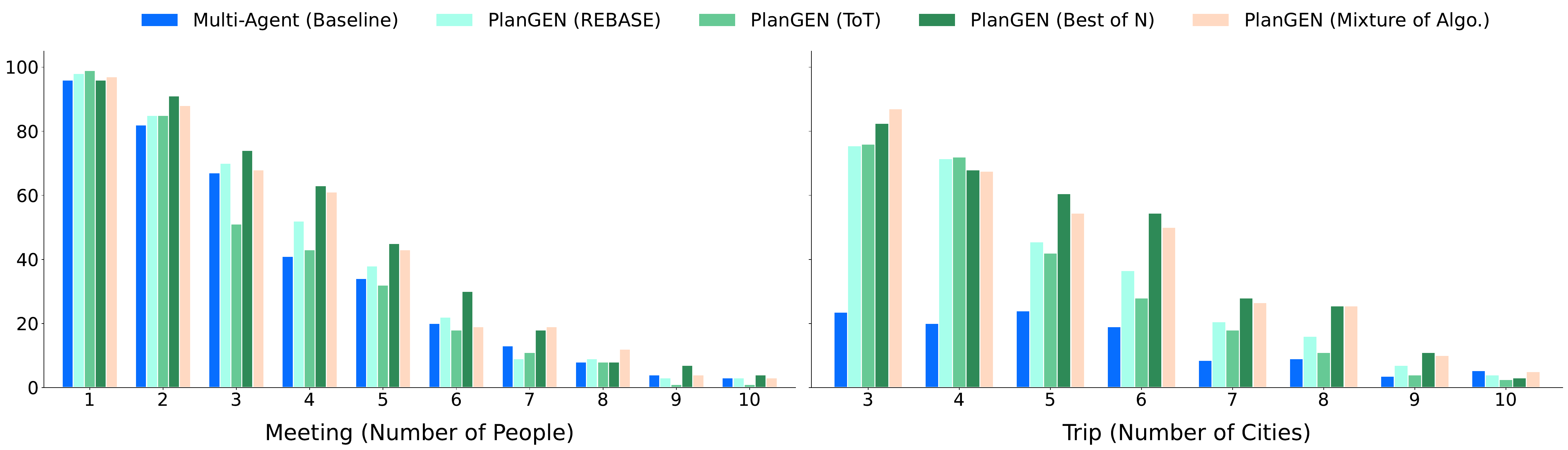}
    \caption{Performance comparison of inference-time algorithms across different complexity levels for meeting and trip planning from NATURAL PLAN}
    \label{fig:np_meet_trip_analysis}
\end{figure}

From Figure \ref{fig:np_meet_trip_analysis}, in the meeting planning, \plangen{} (Best of $\mathcal{N}$) excels in both simple and intermediate problems, whereas a \plangen{} (Mixture of Algo.) performs better for complex problems. The trip planning presents a different trend, where \plangen{} (Best of $\mathcal{N}$) and a \plangen{} (Mixture of Algo.) consistently outperform other approaches across all complexity levels. Nonetheless, in very complex problems for both meeting and trip planning, all algorithms exhibit poor performance.

\begin{wraptable}{r}{0.5\textwidth}
\centering
\vspace{-23mm}
\footnotesize
\resizebox{0.75\linewidth}{!}{
\begin{tabular}{c|cc}
\toprule
\multirow{2}{*}{\textbf{Methods}} & \multicolumn{2}{c}{\textbf{OlympiadBench}}       \\ \cmidrule{2-3} 
                                  & \multicolumn{1}{c|}{\textbf{MATH}} & \textbf{PHY} \\ \midrule
\plangen{} (Best of $\mathcal{N}$) (5)                     & \multicolumn{1}{c|}{53.26}         & 32.63        \\ 
\plangen{} (Best of $\mathcal{N}$) (10)                    & \multicolumn{1}{c|}{54.90}         & 31.36        \\ 
\plangen{} (Best of $\mathcal{N}$) (20)                    & \multicolumn{1}{c|}{53.22}         & 29.38        \\ \midrule
\plangen{} (ToT) (3)                           & \multicolumn{1}{c|}{52.97}         & 31.36        \\ 
\plangen{} (ToT) (5)                           & \multicolumn{1}{c|}{55.20}         & 32.05        \\ 
\plangen{} (ToT) (10)                          & \multicolumn{1}{c|}{55.79}         & 32.52        \\ \midrule
\plangen{} (REBASE) (10)                       & \multicolumn{1}{c|}{54.45}         & 31.78        \\ 
\plangen{} (REBASE) (20)                       & \multicolumn{1}{c|}{54.45}         & 29.37        \\ 
\plangen{} (REBASE) (30)                       & \multicolumn{1}{c|}{55.04}         & 30.28        \\ \bottomrule
\end{tabular}
}
\caption{Performance impact of hyper-parameters on inference-time algorithms in OlympiadBench}
\label{tab:hyperparameters}
\end{wraptable}

\paragraph{Different hyper-parameters of inference-time algorithms vs. their performance}

We conduct a case study on OlympiadBench, where we analyze the impact of varying hyper-parameters on the performance of different inference-time algorithms. The results (Table \ref{tab:hyperparameters}) indicate that while increasing the number of samples (Best of $\mathcal{N}$), steps (ToT), or refinements (REBASE) lead to marginal improvements, the overall differences remain relatively small. Given this, we opted for lower hyper-parameter values across all inference-time algorithms to balance efficiency and performance.

\paragraph{Frequency of inference-time algorithm selection across datasets}

For the \plangen{} (Mixture of Algo.) method, we analyze how frequently each inference-time algorithm (Best of $\mathcal{N}$, ToT, and REBASE) is selected across different datasets. The results (shown in Table \ref{tab:algo_selection}) show that \plangen{} (ToT) is the most frequently chosen algorithm in NATURAL PLAN, OlympiadBench, and GPQA, indicating its effectiveness in these domains. In contrast, for DocFinQA, \plangen{} (Best of $\mathcal{N}$) is the dominant choice, suggesting that its strategy aligns better with financial reasoning tasks. \plangen{} (REBASE) is selected the least across all datasets, implying that its refinements are less favored by the selection mechanism. These findings highlight the dataset-dependent nature of inference-time algorithm effectiveness and the adaptability of the mixture approach in dynamically choosing the most suitable method.

\begin{algorithm*}
\caption{Selection using Modified UCB with LLM-Guided Priors}
\begin{algorithmic}[1]
\State \textbf{Initialize:} $R(a) \gets 0$, $C(a) \gets 1$, $Rec(a) \gets 0$, $F(a) \gets 0$, $D(a) \gets 1$, $T \gets 0$
\State Set $\lambda_{\text{prior}}$, $\alpha_{\text{diversity}}$, $\alpha_{\text{recovery}}$
\State Load LLM-guided priors

\Procedure{SelectAlgorithm}{args}
    \State Compute prior decay: $\lambda_{\text{prior}} \gets \frac{\lambda_{\text{prior}}}{1 + T}$ \Comment{Reduces as trials increase}
    \State Set max exploration term $M \gets 5$
    \State Obtain LLM prior scores: $LLM\_prior \gets \text{LLM\_Guided\_Prior}(args)$
    \State Compute max reward: $R_{\max} \gets \max(R(a))$ (set to 1 if all rewards are 0)

    \For{each algorithm $a$}
        \State Compute normalized reward:
        \[
        \bar{R}_{\text{norm}}(a) \gets \frac{R(a)}{C(a) R_{\max}}
        \]
        \Comment{Scales rewards between 0 and 1 for comparability}

        \State Compute exploration term:
        \[
        E(a) \gets \min\left(\sqrt{\frac{2 \log(T+1)}{C(a)}}, M\right)
        \]
        \Comment{Encourages trying less-used algorithms, capped at $M$}

        \State Compute diversity bonus:
        \[
        D(a) \gets \frac{\alpha_{\text{diversity}}}{C(a) + 1}
        \]
        \Comment{Penalizes frequently used algorithms to encourage variety}

        \State Compute recovery bonus:
        \[
        RecB(a) \gets \alpha_{\text{recovery}} \cdot Rec(a)
        \]
        \Comment{Rewards algorithms that perform well after failures}

        \State Compute final UCB score:
        \[
        UCB(a) \gets \bar{R}_{\text{norm}}(a) + E(a) + \lambda_{\text{prior}} LLM\_prior(a) + D(a) + RecB(a)
        \]
        \Comment{Balances exploitation, exploration, diversity, and recovery}
    \EndFor
    
    \State Select best algorithm:
    \[
    a^* \gets \arg\max_{a} UCB(a)
    \]
    \State \Return $(a^*, UCB(a^*))$
\EndProcedure

\end{algorithmic}
\label{algo:selection}
\end{algorithm*}

\section{Various Examples for Different Components of \plangen{}}
\label{app:examples}

\paragraph{Examples for Constraint Agent}
To illustrate the output of our constraint agent, Table \ref{tab:np_constraints_examples}, Table \ref{tab:gpqa_constraints_examples}, and Table \ref{tab:olympiad_math_constraints} present representative examples of generated constraints. These tables highlight the diverse constraints generated for problem instances of different tasks.

\begin{wraptable}{r}{0.5\textwidth}
\centering
\vspace{-5mm}
\footnotesize
\resizebox{\linewidth}{!}{
\begin{tabular}{l|cccc}
\toprule
Frameworks & NATURAL PLAN & OlympiadBench & GPQA & DocFinQA \\
\midrule
\plangen{} (Best of $\mathcal{N}$) & 19.55\% & 7.09\% & 8.56\% & 81.03\% \\
\plangen{} (ToT) & 68.85\% & 90.09\% & 85.59\% & 12.5\% \\
\plangen{} (REBASE) & 11.6\% & 2.82\% & 5.86\% & 6.47\% \\
\bottomrule
\end{tabular}
}
\caption{Algorithm Selection Frequency by Dataset}
\label{tab:algo_selection}
\end{wraptable}

\paragraph{Example for Verification Agent}
To illustrate the output of our verification agent, Table \ref{tab:np_cal_verification_example} presents representative examples of verification process for NATURAL PLAN (calendar scheduling). This table highlights the how the verification agent verifies the generated plan using constraints.

\paragraph{Examples of Generated Plans}
To demonstrate the plan generation process, Table \ref{tab:np_examples}, Table \ref{tab:gpqa_example}, Table \ref{tab:docfinqa_example}, and Table \ref{tab:olympiad_math_example} present example plans for NATURAL PLAN, GPQA, DocFinQA, and OlympiadBench.  Generated using \plangen{} (Best of $\mathcal{N}$), these tables highlight the varied nature of plans produced across different task types. For GPQA, DocFinQA, and OlympiadBench (i.e., downstream reasoning tasks), the examples additionally illustrate how generated plans are executed to derive final answer. 


\begin{table*}
\centering
\scriptsize
\renewcommand{\arraystretch}{1.2} 

\begin{tabularx}{\textwidth}{>{\raggedright}p{0.1\textwidth}|X|X}
\toprule
\textbf{Benchmark}   & \textbf{Problem Statement}  & \textbf{Generated Constraints} \\ \midrule

Calendar Scheduling & You are an expert at scheduling meetings. You are given a few constraints on the existing schedule of each participant, the meeting duration, and possibly some preferences on the meeting time. Note there exists a solution that works with existing schedule of every participant.

------------

\textbf{TASK:} You need to schedule a meeting for Michelle, Steven and Jerry for one hour between the work hours of 9:00 to 17:00 on Monday.

------------

Here are the existing schedules for everyone during the day:

- Michelle has meetings on Monday during 11:00 to 12:00;

- Steven has blocked their calendar on Monday during 9:00 to 9:30, 11:30 to 12:00, 13:30 to 14:00, 15:30 to 16:00;

 -Jerry has blocked their calendar on Monday during 9:00 to 9:30, 10:00 to 11:00, 11:30 to 12:30, 13:00 to 14:30, 15:30 to 16:00, 16:30 to 17:00;

------------

Find a time that works for everyone's schedule and constraints.

& 

\textbf{Instance-Specific Constraints and Details:}

Participants: Michelle, Steven, Jerry - The meeting must include all three.

------------

Meeting Duration: 1 hour - The meeting requires a contiguous 60-minute slot.

------------

Work Hours: 9:00 to 17:00 (Monday) -  The meeting must start and end within these working hours.  This acts as a global constraint.

------------

Individual Schedules (Existing Meetings/Blocked Times):

- Michelle: Busy from 11:00 to 12:00 on Monday.
    
- Steven: Busy from 9:00 to 9:30, 11:30 to 12:00, 13:30 to 14:00, and 15:30 to 16:00 on Monday.
    
- Jerry: Busy from 9:00 to 9:30, 10:00 to 11:00, 11:30 to 12:30, 13:00 to 14:30, and 15:30 to 17:00 on Monday (note this constraint ends at the end of the workday).
    
------------

While the problem states "Note there exists a solution," this isn't a constraint in itself; it's a guarantee that a feasible solution exists given the constraints. The actual constraints are the schedules, meeting duration, and working hours.

\\ \midrule

Meeting Planning & 
You are visiting San Francisco for the day and want to meet as many friends as possible. Solve the problem by considering various different schedules and picking the best one to optimize your goals.

------------

Travel distances (in minutes):

Bayview to Golden Gate Park: 22.

Golden Gate Park to Bayview: 23.

------------

CONSTRAINTS: You arrive at Bayview at 9:00AM. Barbara will be at Golden Gate Park from 8:00AM to 11:30AM. You'd like to meet Barbara for a minimum of 90 minutes.

------------

Your response should start with 'SOLUTION:'

&  

\textbf{Instance-Specific Constraints and Details:}

Travel Times:
    
    - Bayview to Golden Gate Park: 22 minutes.
    
    - Golden Gate Park to Bayview: 23 minutes (Note: This information is provided but not strictly necessary for this specific problem as the return trip isn't relevant to meeting Barbara).

------------

Locations:
    
    - Starting Location: Bayview
    
    - Meeting Location: Golden Gate Park
    
------------

Scheduled Meeting Times/Participant Availability:
    
    - Barbara is available at Golden Gate Park from 8:00 AM to 11:30 AM.

------------

Arrival Time:
    
    - You arrive at Bayview at 9:00 AM.

------------

Minimum Meeting Duration:
    
    - The meeting with Barbara must be at least 90 minutes long.

------------

Sequence Requirements:
    
    - Travel (Bayview to Golden Gate Park) $\rightarrow$ Wait (if necessary, due to arriving before Barbara's availability ends) $\rightarrow$ Meeting (with Barbara)

------------

Objective: Maximize the number of friends met (in this case, it's simplified to just meeting Barbara, so the objective is to determine IF a meeting is possible given the constraints). 

\\ \bottomrule
\end{tabularx}

\caption{Examples of constraints generated by the constraint agent given problem. These examples illustrate constraint generation for calendar scheduling and meeting planning from NATURAL PLAN}
\label{tab:np_constraints_examples}
\end{table*}

\begin{table*}
\centering
\footnotesize
\begin{tabular}{p{\linewidth}}
\toprule
\textbf{Question:} Two quantum states with energies E1 and E2 have a lifetime of $10^{-9}$ sec and $10^{-8}$ sec, respectively. We want to clearly distinguish these two energy levels. Which one of the following options could be their energy difference so that they can be clearly resolved?

(A) $10^{-11}$ eV

(B) $10^{-8}$ eV

(C) $10^{-9}$ eV

(D) $10^{-4}$ eV \\ \midrule

\textbf{Constraints for Verifying the Energy Difference Solution for Resolvable Quantum States}
\\ \\
This problem revolves around the energy-time uncertainty principle, which dictates the resolvability of energy levels given their lifetimes. To verify a proposed energy difference ($\Delta E = |E_2 - E_1|$) that allows clear resolution, we need constraints encompassing the following:
\\\\
1. Uncertainty Principle Adherence:

\textbf{Primary Constraint:} The product of the uncertainty in energy ($\Delta E$) and the uncertainty in time ($\Delta t$) must satisfy the energy-time uncertainty relation: $\Delta E \Delta t \geq \frac{\hbar}{2}$, where $\hbar$ is the reduced Planck constant.

\textbf{Specific Constraint:} Since we have two states, we must consider the \textit{combined} uncertainty in time. We can approximate this as the average of the individual lifetimes: $\Delta t \approx \frac{\tau_1 + \tau_2}{2}$, where $\tau_1$ and $\tau_2$ are the lifetimes of the two states.

\textbf{Operational Constraint:} For "clear resolution," we typically require $\Delta E$ to be significantly greater than the uncertainty dictated by the uncertainty principle. A common rule of thumb is $\Delta E \geq \frac{\hbar}{\Delta t}$. A stricter constraint might use a larger factor (e.g., $\Delta E \geq \frac{2\hbar}{\Delta t}$) for even clearer resolution.
\\\\
2. Units Consistency:

\textbf{Unit Constraint:} All calculations must use consistent units. Ensure that energy ($\Delta E$), time ($\Delta t$), and $\hbar$ are expressed in compatible units (e.g., Joules and seconds, or eV and seconds). Clearly state the chosen unit system.
\\\\
3. Numerical Accuracy:

\textbf{Precision Constraint:} Calculations should be performed with sufficient numerical precision to avoid rounding errors. Define the acceptable level of precision.

\textbf{Constant Accuracy:} Use a precise value for the reduced Planck constant ($\hbar$).
\\\\
4. Solution Justification:

\textbf{Explanation Constraint:} The solution must not only provide a numerical value for $\Delta E$ but also clearly explain why this value satisfies the constraints and allows for clear resolution. This explanation should reference the uncertainty principle and the chosen resolution criterion.
\\\\
5. Alternative Solutions Consideration (Optional):

\textbf{Exploration Constraint:} Explore whether multiple values of $\Delta E$ could satisfy the constraints. If so, discuss the trade-offs between different solutions.
\\\\
\textbf{Example Implementation of Constraints:}
\\\\
Let's assume we're working in SI units (Joules and seconds).

1. Uncertainty Principle: $\Delta E \geq \frac{\hbar}{[(10^{-9} \text{ s} + 10^{-8} \text{ s})/2]}$, where $\hbar \approx 1.054 \times 10^{-34} \text{ J s}$.

2. Units: All calculations are in Joules and seconds.

3. Numerical Accuracy: Calculations will be performed to at least 4 significant figures.

4. Justification: The solution will explain how the calculated $\Delta E$ is significantly larger than the uncertainty, ensuring clear resolution.

 \\ \bottomrule
\end{tabular}
\caption{Examples of constraints generated by the constraint agent given reasoning question. The example illustrates the constraint generation for solving physics question from GPQA}
\label{tab:gpqa_constraints_examples}
\end{table*}

\begin{table*}
\centering
\footnotesize
\begin{tabular}{p{\linewidth}}
\toprule
\textbf{Question:} For a positive integer $a$, define a sequence of integers $x_{1}, x_{2}, \ldots$ by letting $x_{1}=a$ and $x_{n+1}=2 x_{n}+1$ for $n\geq1$. Let $y_{n}=2^{x_{n}}-1$. Determine the largest possible $k$ such that, for some positive integer $a$, the numbers $y_{1}, \ldots, y_{k}$ are all prime. \\ \midrule

\textbf{Constraints for Verifying the Solution Plan for Prime Sequence Problem}
\\\\
This problem requires a solution plan that likely involves a combination of mathematical reasoning and computational verification. Here's a breakdown of constraints to verify and evaluate such a plan:
\\\\
\textbf{I. Mathematical Reasoning Constraints:}
\\\\
1. \textbf{Correct Recurrence Relation:} The plan must correctly interpret and utilize the given recurrence relations for $x_n$ and $y_n$. Verify that the plan doesn't misinterpret or modify these definitions.

2. \textbf{Mersenne Primes Understanding:} The plan should demonstrate an understanding that $y_n$ are Mersenne numbers (numbers of the form $2^p - 1$). It should acknowledge that for $y_n$ to be prime, $x_n$ must itself be prime (though this is a necessary but not sufficient condition).

3. \textbf{Logical Flow and Justification:} The plan's reasoning should be clear, logical, and well-justified. Each step should follow from the previous ones, with clear explanations for any claims or deductions made. Avoid leaps of logic or unsubstantiated assertions.

4. \textbf{Handling of Base Case ($k=1$):} The plan should explicitly address the trivial case where $k=1$. Since $y_1 = 2^a - 1$, any prime value of 'a' will result in $y_1$ being a potential Mersenne prime. This confirms that $k \ge 1$.

5. \textbf{Strategy for Finding Maximum $k$:} The plan must outline a clear strategy for determining the \textit{largest} possible $k$. This likely involves exploring increasing values of $k$ and checking the primality of the resulting $y_n$ sequence. The plan should explain how it avoids infinite search or unnecessary computations.

6. \textbf{Optimality Argument:} The plan should include an argument (or at least a strategy for developing an argument) for why the found value of $k$ is indeed the \textit{largest} possible. This might involve demonstrating that for $k+1$, no suitable 'a' exists.
\\\\
\textbf{II. Computational Verification Constraints:}
\\\\
1. \textbf{Primality Testing Algorithm:} If the plan involves computational primality testing, it should specify the algorithm used (e.g., probabilistic tests like Miller-Rabin or deterministic tests like AKS). The plan should justify the choice of algorithm based on the size of the numbers being tested and the desired level of certainty.

2. \textbf{Correct Implementation:} If code is involved, it must be correct and free of errors. The plan should include testing procedures to ensure the code accurately implements the chosen primality test and the recurrence relations.

3. \textbf{Computational Resources:} The plan should consider the computational resources required for the primality testing. For large values of $x_n$, testing the primality of $y_n$ can be computationally expensive. The plan should address how it manages these computational challenges.

4. \textbf{Range of 'a' considered:} The plan should specify the range of values for 'a' that will be explored. It should justify this range and explain why exploring values outside this range is unnecessary.

5. \textbf{Termination Condition:} The plan must have a clear termination condition for its computational search. It should explain how it determines when to stop searching for larger values of 'k'.
\\\\
\textbf{III. Presentation and Clarity Constraints:}
\\\\
1. \textbf{Clear and Concise Explanation:} The plan should be presented in a clear, concise, and easy-to-understand manner. Avoid unnecessary jargon or overly complex language.

2. \textbf{Well-Organized Structure:} The plan should be well-organized, with a logical flow of ideas. Use headings, subheadings, and bullet points to improve readability.

3. \textbf{Reproducibility:} The plan should be sufficiently detailed to allow another person to reproduce the results. All assumptions, parameters, and procedures should be clearly stated.

 \\ \bottomrule
\end{tabular}
\caption{Examples of constraints generated by the constraint agent given MATH reasoning problem. The example illustrates the constraint generation for solving MATH problem from OlympiadBench}
\label{tab:olympiad_math_constraints}
\end{table*}

\begin{table*}
\centering
\scriptsize
\begin{tabular}{p{\linewidth}}
\toprule
\textbf{Problem Statement:} You are an expert at scheduling meetings. You are given a few constraints on the existing schedule of each participant, the meeting duration, and possibly some preferences on the meeting time. Note there exists a solution that works with existing schedule of every participant.
\\\\
\textbf{TASK:} You need to schedule a meeting for Michelle, Steven and Jerry for one hour between the work hours of 9:00 to 17:00 on Monday.
\\\\
Here are the existing schedules for everyone during the day:

- Michelle has meetings on Monday during 11:00 to 12:00;

- Steven has blocked their calendar on Monday during 9:00 to 9:30, 11:30 to 12:00, 13:30 to 14:00, 15:30 to 16:00;

 -Jerry has blocked their calendar on Monday during 9:00 to 9:30, 10:00 to 11:00, 11:30 to 12:30, 13:00 to 14:30, 15:30 to 16:00, 16:30 to 17:00;
\\\\
Find a time that works for everyone's schedule and constraints.
\\ \midrule
\textbf{Instance-Specific Constraints and Details:}
\\\\
Participants: Michelle, Steven, Jerry - The meeting must include all three.
\\\\
Meeting Duration: 1 hour - The meeting requires a contiguous 60-minute slot.
\\\\
Work Hours: 9:00 to 17:00 (Monday) -  The meeting must start and end within these working hours.  This acts as a global constraint.
\\\\
Individual Schedules (Existing Meetings/Blocked Times):

    - Michelle: Busy from 11:00 to 12:00 on Monday.
    
    - Steven: Busy from 9:00 to 9:30, 11:30 to 12:00, 13:30 to 14:00, and 15:30 to 16:00 on Monday.
    
    - Jerry: Busy from 9:00 to 9:30, 10:00 to 11:00, 11:30 to 12:30, 13:00 to 14:30, and 15:30 to 17:00 on Monday (note this constraint ends at the end of the workday).
\\\\
While the problem states "Note there exists a solution," this isn't a constraint in itself; it's a guarantee that a feasible solution exists given the constraints. The actual constraints are the schedules, meeting duration, and working hours.
\\ \midrule

\textbf{Generated Plan:}
\\\\
1. Analyze Constraints:

- Participants: Michelle, Steven, Jerry

- Duration: 1 hour

- Work hours: 9:00 - 17:00

- Day: Monday
\\\\
2. Identify Individual Availability:

- Michelle: 9:00-11:00, 12:00-17:00

- Steven: 9:30-11:30, 12:00-13:30, 14:00-15:30, 16:00-17:00

- Jerry: 9:30-10:00, 11:00-11:30, 12:30-13:00, 14:30-15:30, 16:00-16:30
\\\\
3. Determine Common Availability: Looking for 1-hour slots:

- Comparing Michelle and Steven: 9:30-11:00, 12:00-13:30, 14:00-15:30, 16:00-17:00

- Comparing the above with Jerry:  14:30-15:30
\\\\
4. Prioritize Earliest Time Slot: The earliest 1-hour slot is 14:30-15:30
\\\\
5. Ensure Solution Exists:  The slot 14:30-15:30 works for all three.
\\\\
6. Strict Format Requirement:

SOLUTION: Here is the proposed time: Monday, 14:30 - 15:30 \\ \midrule

\textbf{Verification:}
\\\\
The plan demonstrates significant improvement compared to the initial plan.  It correctly identifies individual availabilities and successfully finds the two viable one-hour slots: 9:30-10:30 and 14:30-15:30.  It avoids exploring irrelevant shorter slots.  The plan is well-structured and clearly explains each step.  It's on the verge of providing the correct solution.  However, it has not yet explicitly stated the final answer (the prioritized earliest time slot) which is a key part of the task.  Therefore, while much improved, it doesn't quite merit a top score due to this omission.
\\\\
\textbf{Score: 95}

\\ \bottomrule
\end{tabular}
\caption{Examples of verification generated by the verification agent given generated plan and constraints. This example illustrates verification process for calendar scheduling from NATURAL PLAN}
\label{tab:np_cal_verification_example}
\end{table*}

\begin{table*}
\centering
\scriptsize
\renewcommand{\arraystretch}{1.2} 

\begin{tabularx}{\textwidth}{>{\raggedright}p{0.1\textwidth}|X|X}
\toprule
\textbf{Category}   & \textbf{Problem Statement}  & \textbf{Generated Plan} \\ \midrule

Calendar Scheduling & You are an expert at scheduling meetings. You are given a few constraints on the existing schedule of each participant, the meeting duration, and possibly some preferences on the meeting time. Note there exists a solution that works with existing schedule of every participant.

------------

\textbf{TASK:} You need to schedule a meeting for Michelle, Steven and Jerry for one hour between the work hours of 9:00 to 17:00 on Monday.

------------

Here are the existing schedules for everyone during the day:

- Michelle has meetings on Monday during 11:00 to 12:00;

- Steven has blocked their calendar on Monday during 9:00 to 9:30, 11:30 to 12:00, 13:30 to 14:00, 15:30 to 16:00;

 -Jerry has blocked their calendar on Monday during 9:00 to 9:30, 10:00 to 11:00, 11:30 to 12:30, 13:00 to 14:30, 15:30 to 16:00, 16:30 to 17:00;
 
------------

Find a time that works for everyone's schedule and constraints.

& 

1. Analyze Constraints:

- Participants: Michelle, Steven, Jerry

- Duration: 1 hour

- Work hours: 9:00 - 17:00

- Day: Monday

------------

2. Identify Individual Availability:

- Michelle: 9:00-11:00, 12:00-17:00

- Steven: 9:30-11:30, 12:00-13:30, 14:00-15:30, 16:00-17:00

- Jerry: 9:30-10:00, 11:00-11:30, 12:30-13:00, 14:30-15:30, 16:00-16:30

------------

3. Determine Common Availability: Looking for 1-hour slots:

- Comparing Michelle and Steven: 9:30-11:00, 12:00-13:30, 14:00-15:30, 16:00-17:00

- Comparing the above with Jerry:  14:30-15:30

------------

4. Prioritize Earliest Time Slot: The earliest 1-hour slot is 14:30-15:30

------------

5. Ensure Solution Exists:  The slot 14:30-15:30 works for all three.

------------

6. Strict Format Requirement:

SOLUTION: Here is the proposed time: Monday, 14:30 - 15:30

\\ \midrule

Meeting Planning & 
You are visiting San Francisco for the day and want to meet as many friends as possible. Solve the problem by considering various different schedules and picking the best one to optimize your goals.

------------

Travel distances (in minutes):

Bayview to Golden Gate Park: 22.

Golden Gate Park to Bayview: 23.

------------

CONSTRAINTS: You arrive at Bayview at 9:00AM. Barbara will be at Golden Gate Park from 8:00AM to 11:30AM. You'd like to meet Barbara for a minimum of 90 minutes.

&  

You leave Bayview at 9:00 AM and travel 22 minutes to Golden Gate Park, arriving at 9:22 AM. You meet Barbara from 9:22 AM – 10:52 AM, fulfilling the 90-minute requirement. \\ \midrule

Trip Planning &
You plan to visit 4 European cities for 12 days in total. You only take direct flights to commute between cities. You plan to stay in Budapest for 6 days. You plan to stay in Reykjavik for 5 days. You would like to meet your friends at Reykjavik between day 6 and day 10 to tour together. You want to spend 2 days in Milan. You are going to attend a wedding in Milan between day 10 and day 11. You would like to visit Porto for 2 days.

------------

Here are the cities that have direct flights: 

Budapest and Reykjavik, Reykjavik and Milan, Budapest and Milan, Milan and Porto.

------------

Find a trip plan of visiting the cities for 12 days by taking direct flights to commute between them. & 
Day 1-6: Arriving in Budapest and visit Budapest for 6 days.

Day 6: Fly from Budapest to Reykjavik.

Day 6-10: Visit Reykjavik for 5 days.

Day 10: Fly from Reykjavik to Milan.

Day 10-11: Visit Milan for 2 days.

Day 11: Fly from Milan to Porto.

Day 11-12: Visit Porto for 2 days.

\\ \bottomrule
\end{tabularx}

\caption{Examples of generated plans by \plangen{} (Best of $\mathcal{N}$) given problem statement. These examples illustrate the plan generation for all three task categories from NATURAL PLAN}
\label{tab:np_examples}
\end{table*}

\begin{table*}
\centering
\footnotesize
\begin{tabular}{p{\linewidth}}
\toprule
\textbf{Question:} Two quantum states with energies E1 and E2 have a lifetime of $10^{-9}$ sec and $10^{-8}$ sec, respectively. We want to clearly distinguish these two energy levels. Which one of the following options could be their energy difference so that they can be clearly resolved? \\ \midrule
\textbf{Plan to solve the problem:}\\ \\ 1. Identify the relevant physics principle: This problem involves the energy-time uncertainty principle, which states that the product of the uncertainty in energy ($\Delta E$) and the uncertainty in time ($\Delta t$) is greater than or equal to h-bar/2 ($\hbar/2$). The uncertainty in time is related to the lifetime of the states. To clearly distinguish the energy levels, the energy difference between them must be greater than the uncertainty in their energies.\\ \\ 2. Relate lifetimes to time uncertainty: The lifetimes given ($10^{-9}$ s and $10^{-8}$ s) represent the uncertainties in the time for each state ($\Delta t_1$ and $\Delta t_2$). We need to consider which lifetime is more restrictive for resolving the energy levels. Since we want to *clearly* distinguish them, we should use the *smaller* lifetime as the relevant $\Delta t$. This is because the shorter lifetime leads to a larger energy uncertainty, making the levels harder to resolve.\\ \\ 3. Calculate the minimum energy difference: Use the energy-time uncertainty principle ($\Delta E * \Delta t \geq \hbar/2$) with the smaller lifetime to calculate the minimum resolvable energy difference ($\Delta E$). We can approximate this by setting $\Delta E * \Delta t \approx \hbar/2$. We'll need the value of $\hbar$ (Planck's constant divided by $2\pi$) in eV·s.\\ \\ 4. Compare the calculated $\Delta E$ with the choices: Compare the calculated minimum resolvable energy difference ($\Delta E$) with the given options (A) through (D). The correct answer will be the option that is *greater* than the calculated $\Delta E$, ensuring that the energy difference is large enough to resolve the two levels.\\ \\ 5. Select the correct answer: Based on the comparison in step 4, choose the option that represents an energy difference greater than the minimum resolvable energy difference. \\ \midrule
\textbf{Plan Execution:}\\ \\ 1. Identify the relevant physics principle: The energy-time uncertainty principle ($\Delta E * \Delta t \geq \hbar/2$) governs this situation.\\ \\ 2. Relate lifetimes to time uncertainty: The shorter lifetime, $10^{-9}$ s, is the relevant $\Delta t$ because it leads to a larger energy uncertainty, making resolution more difficult. So, $\Delta t = 10^{-9}$ s.\\ \\ 3. Calculate the minimum energy difference:\\ We have $\Delta E * \Delta t = \hbar/2$. We need $\hbar$ in eV·s. $\hbar = 6.582 \times 10^{-16}$ eV·s.\\ $\Delta E = (\hbar/2) / \Delta t$\\ $\Delta E = (6.582 \times 10^{-16} \text{ eV·s} / 2) / (10^{-9} \text{ s})$\\ $\Delta E = 3.291 \times 10^{-7}$ eV\\ This is approximately $3 \times 10^{-7}$ eV.\\ \\ 4. Compare the calculated $\Delta E$ with the choices: We need an energy difference greater than $3 \times 10^{-7}$ eV to clearly resolve the levels.\\ (A) $10^{-11}$ eV - Too small\\ (B) $10^{-8}$ eV - Too small\\ (C) $10^{-9}$ eV - Too small\\ (D) $10^{-4}$ eV - Large enough\\ \\ 5. Select the correct answer: Option (D) is the only one significantly larger than the calculated minimum energy difference.\\ \\ Answer: D \\ \bottomrule
\end{tabular}
\caption{Examples of generated plans by \plangen{} (Best of $\mathcal{N}$) given physics problem. The example illustrates the plan generation and it's execution for physics problem from GPQA}
\label{tab:gpqa_example}
\end{table*}

\begin{table*}
\centering
\footnotesize
\begin{tabular}{p{\linewidth}}
\toprule
\textless{}Entire Document\textgreater\\ \\ \textbf{Question:} What is the percentage change in cash flow hedges in 2011 compare to the 2010? \\ \midrule
\textbf{Plan to solve the problem:}\\ \\ 1. Locate the information on cash flow hedges: Search the document for "cash flow hedges" or similar terms related to derivative instruments used for hedging cash flows.  Focus on sections discussing derivative instruments, hedging activities, and related notional amounts.\\ \\ 2. Identify the notional amounts for 2011 and 2010: Once the relevant section is found (Note 2i), extract the notional amounts of the cash flow hedges specifically for fiscal year 2011 and fiscal year 2010.\\ \\ 3. Calculate the percentage change: Use the formula for percentage change: {[}(Notional Amount 2011 - Notional Amount 2010) / Notional Amount 2010{]} * 100.  This will provide the percentage change in cash flow hedges from 2010 to 2011. \\ \midrule
\textbf{Plan Execution:}\\ \\ 1. Locate the information on cash flow hedges: The information on cash flow hedges was located in a section discussing derivative instruments and hedging activities. Specifically, it was found within a sentence detailing the notional amounts of various hedging instruments.\\ \\    Code (if needed):\\    ```python\\    \# No code needed for this step.\\    ```\\    Result: Information on cash flow hedges located successfully.\\ \\ 2. Identify the notional amounts for 2011 and 2010: The notional amounts for cash flow hedges were identified as \$153.7 million for 2011 and \$139.9 million for 2010.\\ \\    Code (if needed):\\    ```python\\    \# No code needed for this step.\\    ```\\    Result: 2011 notional amount: \$153.7 million; 2010 notional amount: \$139.9 million.\\ \\ 3. Calculate the percentage change:\\ \\    Code (if needed):\\    ```python\\    notional\_2011 = 153.7\\    notional\_2010 = 139.9\\    percentage\_change = ((notional\_2011 - notional\_2010) / notional\_2010) * 100\\    rounded\_percentage\_change = round(percentage\_change, 1)\\    print(f"\{rounded\_percentage\_change\}\%")\\    ```\\ \\ Answer: 9.9\% \\ \bottomrule
\end{tabular}
\caption{Examples of generated plans by \plangen{} (Best of $\mathcal{N}$) given financial reasoning problem. The example illustrates the plan generation and it's execution for problem from DocFinQA}
\label{tab:docfinqa_example}
\end{table*}

\begin{table*}
\centering
\scriptsize
\begin{tabular}{p{\linewidth}}
\toprule
\textbf{Question:} For a positive integer $a$, define a sequence of integers $x_{1}, x_{2}, \ldots$ by letting $x_{1}=a$ and $x_{n+1}=2 x_{n}+1$ for $n\geq1$. Let $y_{n}=2^{x_{n}}-1$. Determine the largest possible $k$ such that, for some positive integer $a$, the numbers $y_{1}, \ldots, y_{k}$ are all prime. \\ \midrule
\textbf{Plan for Solving the IMO Problem:}\\ 1. Problem Understanding: The problem defines a recursive sequence `x\_n' based on an initial value `a', and a related sequence `y\_n' based on `x\_n'. The goal is to find the largest `k' such that `y\_1', `y\_2', ..., `y\_k' can all be prime numbers for some choice of `a'.\\2. Key Information and Variables:\\ `a': A positive integer, the starting value for the sequence `x\_n'. This is the variable we need to choose optimally.\\ `x\_n': Defined recursively as `x\_1 = a' and `x\_(n+1) = 2x\_n + 1'.\\ `y\_n': Defined as `y\_n = 2\textasciicircum{}(x\_n) - 1'.  This is the sequence we need to check for primality.\\ `k': The length of the subsequence of `y\_n' consisting only of primes. We want to maximize this.\\3. Problem Type and Method: This is a number theory problem involving sequences, primality, and recursion. The method will involve a combination of:\\ Analyzing the structure of `x\_n' and 'y\_n':  Finding a closed-form expression for 'x\_n' in terms of 'a' and `n` will be crucial. This will help us understand the structure of `y\_n'.\\ Primality testing/properties: We'll need to leverage known properties of prime numbers, particularly those related to Mersenne primes (primes of the form 2\textasciicircum{}p - 1, where p is prime).\\ Case analysis/exploration: We might need to explore small values of `a' and `k' to identify patterns and formulate conjectures.\\ Proof by contradiction:  To establish the largest possible `k`, we may need to assume a larger value is possible and derive a contradiction.\\ 4. Solution Process:\\ Step 1: Derive a closed-form expression for `x\_n`.  The recursive definition suggests a linear recurrence relation. We can solve it to express `x\_n` directly in terms of `a` and `n`.\\ Step 2: Substitute `x\_n` into the definition of `y\_n`. This gives us a formula for `y\_n` in terms of `a` and `n`.\\ Step 3: Analyze the primality conditions for `y\_n`. Given that `y\_n` has the form 2\textasciicircum{}m - 1,  we know that for `y\_n` to be prime, `m` (which is `x\_n`) must itself be prime.  This suggests focusing on values of `a` that lead to prime values of `x\_n`.\\ Step 4: Explore small values of `a` and `k`. This helps to identify patterns and limitations on the possible values of `k`.\\ Step 5:  Investigate divisibility properties. Consider the factors of `y\_n` and how they might relate to previous terms in the sequence.  Look for patterns in the factors that might prevent multiple consecutive `y\_n` from being prime.\\ Step 6: Formulate a conjecture for the maximum `k`. Based on the observations from the previous steps, propose a value for the maximum `k`.\\ Step 7: Prove the conjecture. Use a proof by contradiction. Assume a larger `k` is possible and demonstrate that this leads to a contradiction based on the divisibility properties of `y\_n`. This will establish the proposed value of `k` as the maximum. \\ \midrule
\textbf{Plan Execution:}\\ 1. Closed-form for $x_n$: The recurrence relation $x_{n+1} = 2x_n + 1$ with $x_1 = a$ can be solved as follows:\\ $x_2 = 2a + 1$\\ $x_3 = 2(2a+1) + 1 = 4a + 3 = 2^2a + 2^2 - 1$\\ $x_4 = 2(4a+3) + 1 = 8a + 7 = 2^3a + 2^3 - 1$\\ By induction, we can show that $x_n = 2^{n-1}a + 2^{n-1} - 1 = 2^{n-1}(a+1) - 1$.\\ 2. $y_n$ in terms of $a$ and $n$: Substituting the expression for $x_n$ into $y_n = 2^{x_n} - 1$, we get $y_n = 2^{2^{n-1}(a+1) - 1} - 1$.\\ 3. Primality of $y_n$: For $y_n$ to be prime, $x_n$ must be prime.  Let $x_n = p_n$, where $p_n$ is a prime number.\\ 4. Exploring small values:\\ If $a=1$, $x_1 = 1$, $y_1 = 2^1 - 1 = 1$, which is not prime.\\ If $a=2$, $x_1 = 2$, $y_1 = 2^2 - 1 = 3$, which is prime.\\ $x_2 = 2(2) + 1 = 5$, $y_2 = 2^5 - 1 = 31$, which is prime.\\ $x_3 = 2(5) + 1 = 11$, $y_3 = 2^{11} - 1 = 2047 = 23 \cdot 89$, which is not prime.  So, for $a=2$, $k=2$.\\ 5. Divisibility:\\ If $x_1 = a$ is prime, say $a=p$, then $x_2 = 2p+1$.  If $x_2$ is also prime, then $y_1$ and $y_2$ are potentially Mersenne primes.  However, $x_3 = 2(2p+1)+1 = 4p+3$.  If $p=2$, $x_3 = 11$, and $y_3$ is not prime.  If $p=3$, $x_2 = 7$, $x_3 = 15$, which is not prime.  If $p=5$, $x_2 = 11$, $x_3 = 23$, $x_4 = 47$, $x_5 = 95$, which is not prime.\\ \\ If $k=3$, we need $x_1=a$, $x_2=2a+1$, and $x_3=4a+3$ to all be prime.  If $a=2$, $x_1=2$, $x_2=5$, $x_3=11$.  But $y_3 = 2^{11}-1 = 2047 = 23 \cdot 89$, which is not prime.\\ \\ 6. Conjecture: The largest possible $k$ is 2.\\ 7. Proof: We have shown that for $a=2$, $y_1$ and $y_2$ are prime.  We need to show that $y_3$ cannot be prime.  $x_3 = 4a+3$.  If $a \equiv 2 \pmod{3}$, then $4a+3 \equiv 4(2)+3 \equiv 11 \equiv 2 \pmod{3}$.  If $a \equiv 1 \pmod{3}$, then $4a+3 \equiv 4(1)+3 \equiv 7 \equiv 1 \pmod{3}$.  If $a \equiv 0 \pmod{3}$, then $4a+3 \equiv 3 \equiv 0 \pmod{3}$.  So, if $a>3$ and $a \equiv 0 \pmod{3}$, then $x_3$ is divisible by 3 and greater than 3, so it is not prime.  If $a=3$, $x_3 = 15$, not prime.  If $a=2$, $x_3 = 11$, and $y_3 = 2047 = 23 \cdot 89$.\\ \\ So the final answer is $\boxed{2}$. \\ \bottomrule
\end{tabular}
\caption{Examples of generated plans by \plangen{} (Best of $\mathcal{N}$) given maths problem. The example illustrates the plan generation and it's execution for problem from OlympiadBench (MATH)}
\label{tab:olympiad_math_example}
\end{table*}

\end{document}